\pdfoutput=1

\documentclass[11pt]{article}

\usepackage[final]{acl}

\usepackage{times}
\usepackage{latexsym}

\usepackage{url}
\usepackage{todonotes} 
\usepackage{subcaption}
\usepackage{amssymb}
\usepackage{enumitem}
\usepackage{array}
\usepackage{longtable}
\usepackage{makecell}
\usepackage{xspace}
\usepackage{xcolor,colortbl}
\usepackage{tcolorbox}
\usepackage{arydshln}
\usepackage{adjustbox}

\usepackage{pgfplots}
\pgfplotsset{width=1.0\columnwidth}
\usepackage{multirow}
\usepackage{makecell}
\usepackage{pifont}
\usepackage{bbm}
\usepackage{rotating}
\usepackage{amsmath}  
\usepackage{tablefootnote}
\usepackage{soul}
\usepackage{booktabs}
\usepackage{tikz}
\usepackage[tikz]{bclogo}
\usepackage{pgfplotstable}
\usepackage{mdframed}
\usepackage{graphicx}
\usepackage{amsmath}
\usepackage{algorithm}
\usepackage{algpseudocode}
\usepackage{wrapfig}
\usepackage[fixed]{fontawesome5}

\usepackage[T1]{fontenc}

\usepackage[utf8]{inputenc}

\usepackage{microtype}

\usepackage{inconsolata}

\newcommand{\planicon}[0]{{\small{\faList*[regular]}}}
\newcommand{\groundicon}[0]{{\small{\faAtom}}}
\newcommand{\exeicon}[0]{{\small{\faTools}}}

\newcommand{\methodname}[0]{\textsc{Lumos}}
\newcommand{\methodnamei}[0]{\textsc{Lumos-I}}
\newcommand{\methodnameo}[0]{\textsc{Lumos-O}}

\definecolor{darkblue}{rgb}{0, 0, 0.5}
\hypersetup{citecolor=darkblue}

\definecolor{beaublue}{rgb}{0.74, 0.83, 0.9}
\DeclareRobustCommand{\hlcyan}[1]{{\sethlcolor{beaublue}\hl{#1}}}
\DeclareRobustCommand{\hlpink}[1]{{\sethlcolor{pink}\hl{#1}}}

\newmdenv[
  backgroundcolor=purple!10,
  skipabove=1em,
  skipbelow=0em,
  leftline=true,
  topline=false,
  bottomline=false,
  rightline=false,
  linecolor=purple!88,
  linewidth=4pt
]{githubquote}

\title{
{\vspace{-2em}{\small \hfill ACL 2024 Main Conference}\\
\vspace*{.2in}}
\includegraphics[width=0.8em, trim=0 0 0 0, clip]{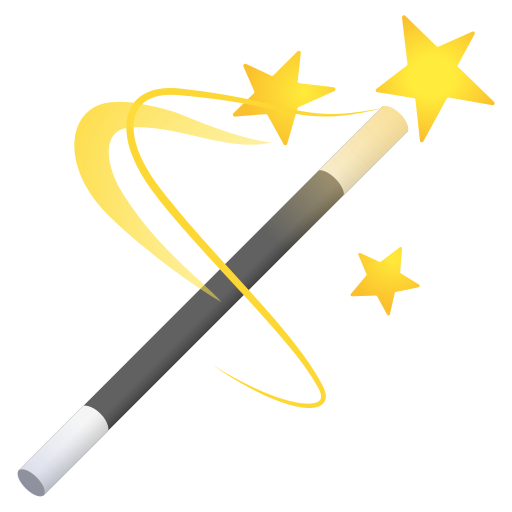} Agent \methodname{}: \\
Unified and Modular Training for Open-Source Language Agents
}

\author{
    \textbf{Da Yin}$^\heartsuit$\thanks{\ Work was done during Da's internship at AI2. Code: \url{https://github.com/allenai/lumos}. Models \& Data: \url{https://huggingface.co/ai2lumos}} \quad
    \textbf{Faeze Brahman}$^\spadesuit$ \quad 
    \textbf{Abhilasha Ravichander}$^\spadesuit$ \quad 
    \textbf{Khyathi Chandu}$^\spadesuit$ \\
    \textbf{Kai-Wei Chang}$^\heartsuit$ \quad 
    \textbf{Yejin Choi}$^{\spadesuit\diamondsuit}$ \quad 
    \textbf{Bill Yuchen Lin}$^\spadesuit$ \\
    [8pt]
    $^\spadesuit$Allen Institute for AI \\
    $^\heartsuit$UCLA \quad
    $^\diamondsuit$University of Washington \\[3pt]
    {\url{https://allenai.github.io/lumos/}}
    \\
    [5pt]
    {   
        \texttt{da.yin@cs.ucla.edu} \quad \texttt{yuchenl@allenai.org} 
    }
}

\begin{document}

\maketitle

\begin{abstract}

  Closed-source agents suffer from several issues such as a lack of affordability, transparency, and reproducibility, particularly on complex interactive tasks. This motivates the development of open-source alternatives. We introduce \includegraphics[width=1em, trim=0 0 0 0, clip]{figure/lumos_icon.png} \methodname{}, one of the first frameworks for training open-source LLM-based agents. \methodname{} features a learnable, unified and modular architecture with a planning module that learns high-level subgoal generation, and a grounding module trained to translate these into the actions using various tools in the execution module. The design allows for modular upgrades and wider applicability to diverse interactive tasks.
  To foster generalizable agent learning, we collect large-scale, unified, and high-quality training annotations derived from diverse ground-truth reasoning rationales across various complex interactive tasks. On 9 datasets, \methodname{} exhibits several key advantages: (1) \methodname{} excels multiple larger open-source agents on the held-out datasets (unused for training) for each task type. \methodname{} even surpasses GPT agents on QA and web tasks; (2) \methodname{} outperforms open-source agents produced by chain-of-thoughts and unmodularized integrated training; and (3) \methodname{} effectively generalizes to unseen tasks, outperforming 33B-scale agents and domain-specific agents. 
\end{abstract} 

\section{Introduction}

\label{sec:intro}
Language agents execute actions and interact with external tools or environments, in service of a goal. They have evolved into crucial elements of AI systems targeted at solving complex interactive tasks. These tasks often require agents to perform long-horizon planning and interactive reasoning, and can range from QA~\citep{yang-etal-2018-hotpotqa, geva-etal-2021-aristotle}, to web tasks~\citep{deng2023mind2web,zhou2023webarena}, math \citep{cobbe2021training}, and multimodal reasoning~\citep{schwenk2022okvqa,lu2022learn}.

\begin{figure}[t]
\centering
\includegraphics[width=\columnwidth, trim=0 160 170 0, clip]{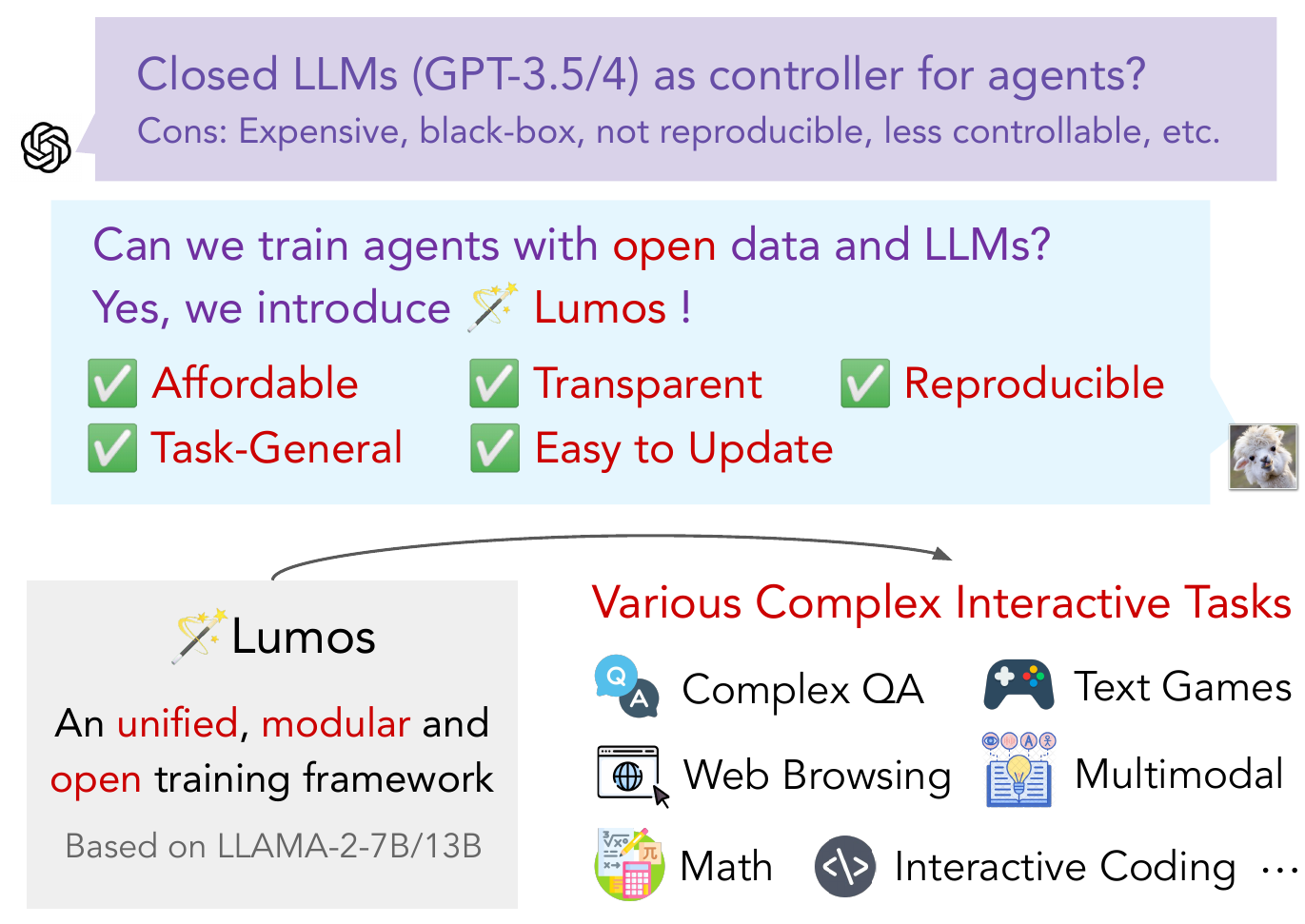}
\caption{\methodname{} is an unified, modular and open-source agent training framework that enables effective cross-task generalization while being easy to be updated. It also has advantages against closed-source agents from affordability, transparency and reproducibility aspects.}
\label{fig:intro-formulations-tease}
\vspace{-9pt}
\end{figure}

Prior agent frameworks~\citep{yao2022react,shinn2023reflexion,Lin2023SwiftSageAG,lu2023chameleon,liu2023bolaa} have primarily relied on closed-source large language model (LLM) APIs such as GPT-4 and ChatGPT~\citep{OpenAI2023GPT4TR,chatgpt_paper}. Though powerful, they can be prohibitively expensive, particularly for tasks with long contexts such as web tasks (which include encoding long HTML code). 
Furthermore, the lack of transparency in closed-source LLMs hinders scientific understanding of their architectures and effectiveness, and provides limited reproducibility, and controllability over their behavior. We argue that over reliance on closed-source LLM-based agents is not conducive to the growth of research on language agents. 

\begin{figure*}[htbp]
\centering
\vspace{-3em}
\includegraphics[width=\textwidth, trim=0 230 0 0, clip]{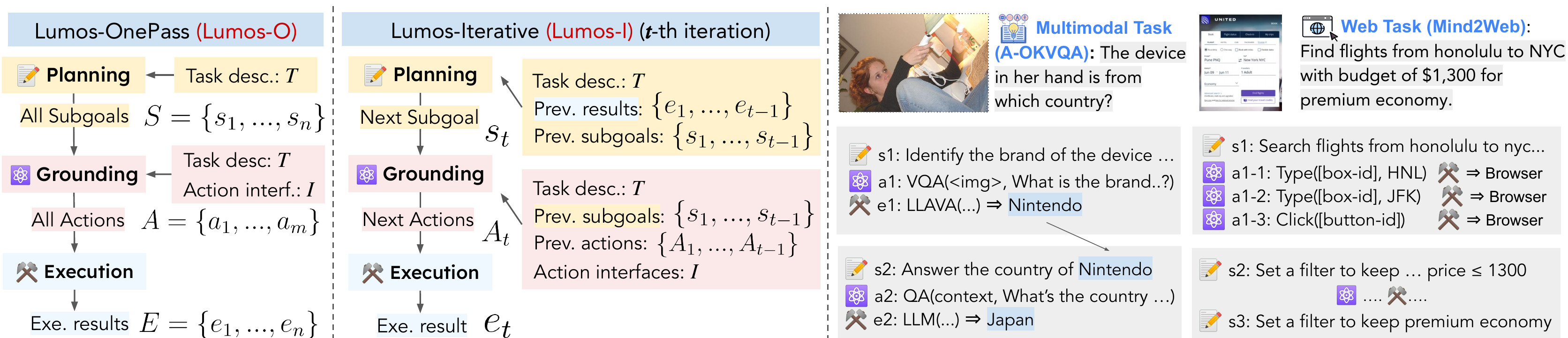}
\caption{Overall framework of \methodname{}. \methodname{} are trained with 56K high-quality training annotations. We propose two agent training and inference formulations, \methodnameo{} (\S\ref{sec:formulations_OnePass}) and \methodnamei{} (\S\ref{sec:formulations_iterative}). \methodnameo{} is an efficient formulation that enables one-pass inference; \methodnamei{} is an adaptive formulation that help agents flexibly plan based on the execution feedback. We showcase two \methodnamei{} running examples in A-OKVQA and Mind2Web.}
\label{fig:intro-formulations}
\vspace{-3pt}
\end{figure*}

In this paper, we propose \includegraphics[width=1em, trim=0 0 0 0, clip]{figure/lumos_icon.png} \methodname{},
a generalizable \textbf{L}anguage agent framework via \textbf{U}nified, \textbf{M}odular, and \textbf{O}pen \textbf{S}ource training. \methodname{} employs a unified and modular architecture broadly applicable to complex interactive tasks: a planning module\planicon{}, a grounding module\groundicon{}, and an execution module\exeicon{}.
The planning module learns to decompose diverse complex tasks into a sequence of high-level subgoals. The grounding module is trained to communicate with the planning module and translate its generated subgoals into the actions that can be executed through a collection of tools in the execution module. \methodname{} design allows for easy module upgrades to enhance new task planning, novel action grounding and supplementing new tools, without impacting the others. 
To tackle the tasks through the agent modules, we propose two interaction formulations for implementing the language agents, \methodname{}-OnePass (\methodnameo{}) and \methodname{}-Iterative (\methodnamei{}). 
Outlined in Fig.~\ref{fig:intro-formulations}, \methodnameo{} is an efficient formulation that generates \textit{all} the subgoals and actions through a \textit{single} inference call, accomplishing the task in a one-pass manner. \methodnamei{} is an iterative formulation that generates one subgoal at a time based on its previous execution results and environment updates, thereby enabling an adaptive agent.

In addition, \methodname{} utilizes a unified data format that encompasses multiple task types, thereby enabling the proposed agent framework to conveniently support a range of interactive tasks. These include, but are not limited to: question answering, mathematics, coding, web browsing, multimodal reasoning, and text games. 
To obtain high-quality annotations for training \methodname{}, we leverage the ground-truth rationales in existing benchmarks across various task types, and convert them into a unified format (\S\ref{sec:annotations}). This conversion is achieved with the aid of strong LLMs, ensuring that the converted annotations adhere to a universally applicable format consistent with our modular design.
Our proposed annotation conversion method results in around 56K multi-task multi-domain agent training annotations, one of the largest open-source resources for agent fine-tuning. The training annotations could serve as a resource for universally enhancing any open-source LLMs with agent capabilities.

Our evaluation demonstrates that \methodname{} provides improved or comparable performance with GPT-based or larger open-source agents across various complex interactive tasks that are commonly used for agent evaluation, including QA, web, math, and multimodal tasks.
We summarize our contributions and results as follows:


\paragraph{General Agent Framework with High-Quality Data.} We introduce an open-source agent learning framework that trains LLMs with unified data, aimed at unifying complex interactive tasks and enhancing generalization on unseen tasks with new environments and actions. We hope our framework and annotations can facilitate future research in developing open-source language agents.

\paragraph{Competitive Performance Across Tasks and Agent Training Formulations.} \methodname{} outperforms a great number of open-source agents on the \methodname{} held-out datasets unused in \methodname{} training data across the four training task types. 
\methodname{} even surpasses GPT-based agents in web and QA tasks. Specifically, \methodname{} shows a 5.0\% enhancement over GPT-4 on Mind2Web, and 4.1\% and 3.5\% LLM accuracy\footnote{A metric defined in \citet{xu2023rewoo} to identify the semantic equivalence between predictions and gold answers.} improvement on HotpotQA over the ReAct and ReWOO agents fully based on GPT-3.5-turbo, respectively. Furthermore, we observe that \methodname{} training formulations outperform other potential agent training methods, such as chain-of-thoughts and integrated training, which instruct a single module to both plan and ground. 

\paragraph{Cross-Task Generalization.} We evaluate \methodname{} on two unseen tasks, WebShop~\citep{yao2022webshop}, a text game for online shopping, and InterCode$_{\mathrm{SQL}}$~\cite{yang2023intercode}, an interactive code generation task. \methodname{} even surpasses 30B-scale agents, especially by nearly 20 reward points on WebShop. \methodname{} also delivers a consistent reward improvement over domain-specific agents. This suggests that \methodname{} can generalize across tasks, hinting at potential benefits for a wide spectrum of language agent applications.


\section{\includegraphics[width=1em, trim=0 0 0 0, clip]{figure/lumos_icon.png} \methodname{}: A Modular Open-Source LLM-Based Agent Framework}
    
    

\label{sec:formulations}
We introduce the overall design and two formulations for developing agents within this framework.






\subsection{\methodname{} Agent Architecture}
\label{sec:formulations_modules}
For various complex interactive tasks, a common solution would include: (1) decomposing the task into a series of subgoals, (2) converting subgoals into concrete actions, (3) executing those actions. This process corresponds to the planning, grounding, and execution modules in our framework.



\paragraph{\planicon{}~{Planning Module} (PM).}
This {module} is designed to dissect a complex task into a series of high-level \texttt{subgoals}, expressed in natural language. For example, a multimodal question such as ``\texttt{The device in her hand is from which country?}'' necessitates two subgoals: 
(1) Identify the brand of the device in her hand; 
(2) Answer the country of the device brand.
The devised subgoals assist in breaking down a complex task into low-level actions in an interpretable and tool-agnostic manner. The planning module is designed for easy debugging and learning new task planning, without affecting other modules. 

\paragraph{\groundicon{}{Grounding Module} (GM).}
This module transforms the high-level subgoals produced by the PM into low-level executable \texttt{actions}. For instance, the GM translates the subgoal, ``\texttt{Query the living period of Lowell Sherman},'' into one or more actions, such as \texttt{KnowledgeQuery(Lowell Sherman)} and \texttt{QA([R2], Query:``What is the living period of Lowell Sherman?'')}. Here, \texttt{R2} refers to the previously retrieved knowledge that may be helpful in answering the query. The grounding module can be easily customized to learn new actions without impacting the planning module.

\paragraph{\exeicon{}~{Execution Module (EM).}}
The Execution Module (EM) is a program that implements the actions generated by the grounding module and gets \texttt{execution results}. It deploys a variety of off-the-shelf tools, including APIs, neural models, and virtual simulators. For instance, the execution module could call the Wikipedia or Google Search APIs to accomplish the \texttt{KnowledgeQuery} action.



The main characteristic of the \methodname{} framework is the interaction among the three modules. We propose two formulations promoting the communication: \methodname{}-OnePass (\methodnameo{}) and \methodname{}-Iterative (\methodnamei{}). 


\subsection{\methodname{}-OnePass (\methodnameo{})}
\label{sec:formulations_OnePass}


The \methodname{}-OnePass (\methodnameo{}) formulation is an efficient method that generates all subgoals and grounded actions at once (efficiency study in App.~\ref{app:efficiency}). As depicted in Fig.~\ref{fig:intro-formulations}, this formulation uses the planning module to generate all $n$ subgoals in a single inference call. We then pass all the generated subgoals to the grounding module, which translates them into a sequence of $m$ low-level actions. Note that in addition to the task description and subgoals, we also provide action interfaces $I$ to the grounding module as inputs. These action interfaces (e.g., \texttt{``VQA(Image\_Context, Query): Given the image context, answer the query.''}) define the functionalities of actions and their valid arguments, guiding the grounding module to produce executable actions.
Lastly, for example, the grounding module can produce all the corresponding actions, from \texttt{VQA([IMG], Question: What's the brand of the device in her hand?)} to the final one \texttt{QA(..., Question: What's the country of ...?)}.

Formally, the overall planning and grounding process of \methodnameo{} is illustrated in Fig.~\ref{fig:intro-formulations}. In the planning phase, the task description $T$ is input into the planning module. This generates an output series of subgoals, expressed as $S=\pi_{plan}(T)=\{s_1, ... , s_n\}$, where $\pi_{plan}$ is the parameters of trained planning module. Grounded actions are obtained via $A = \pi_{ground}(T, I, S)$, with reliance on the task description, action interfaces $I=\{i_1, ... , i_k\}$, and the generated subgoals $S$. $\pi_{ground}$ represents the parameters of the grounding module. We take the last execution result $e_n$ as the final inference result for the given task.

\subsection{\methodname{}-Iterative (\methodnamei{})}
\label{sec:formulations_iterative}
\methodname{}-Iterative (\methodnamei{}) is a formulation that generates one subgoal and its corresponding executable actions in each iteration. When generating the $t$-th subgoal, the planning module requires the previous planned subgoals and the execution results of their grounded actions as input. The execution results assist the planning module to be aware of the environmental change and decide next subgoal according to the up-to-date environments. 


Take the VQA question ``\texttt{The device
in her hand is from which country?}'' in Fig.~\ref{fig:intro-formulations} as an example. In the first iteration, the planning module will produce 
``\texttt{Subgoal 1: Identify the brand of the device in her hand}''. This subgoal is passed to the grounding module to generate the query actions, and obtain the executed results \hlcyan{\texttt{Nintendo}}. The planning module then takes \hlcyan{\texttt{Nintendo}} along with the prior planning context as input to generate the next subgoal ``\texttt{Subgoal 2: Answer the country of Nintendo}''.
Planning upon the latest execution results would mitigate the risk of introducing a non-existent object in the environment or a random entity during the reasoning process (a case study in App.~\ref{app:efficiency}).

We demonstrate a single iteration of planning and grounding process of \methodnamei{} in Fig. \ref{fig:intro-formulations}. To plan the $t$-th subgoal, we input the 1) task description $T$, 2) prior subgoals $\{s_1, ... , s_{t-1}\}$, and 3) their executed results $\{e_1, ... , e_{t-1}\}$ into the planning module. We concatenate them in the order of $T, s_1, e_1, ... , s_{t-1}, e_{t-1}$ where the most recent subgoals and their results are placed in the end, as they have higher influence for planning $t$-th subgoal. The output would be the $t$-th subgoal, $s_t=\pi_{plan}(T, s_1, e_1, ... , s_{t-1}, e_{t-1})$. After then, the $t$-th subgoal will be directly incorporated into grounding module together with the prior grounding history and action interface $I$ to generate the corresponding actions, 
$A_t=\pi_{ground}(T, I, s_1, A_1, ... , s_{t-1}, A_{t-1}, s_t)$. Note that $A_t$ is an executable action \emph{list}, as the high-level subgoal might be decomposed into multiple low-level actions. We finally put the low-level actions $A_t$ into execution module. The final execution result $e_t$ can be sent back for planning $(t+1)$-th subgoal.

\section{Learning to Plan \& Ground with Open-Source LLMs}
\label{sec:annotations}

To guide planning and grounding modules to generate subgoals and valid low-level actions under our specified action interfaces, we fine-tune the two modules to produce the expected outputs.

Training the modules requires the supervisions consisting of high-quality tasks, subgoals, and low-level actions. To equip smaller LLMs with instruction-following ability, prior works leverage methods such as Self-Instruct~\citep{wang-etal-2023-self-instruct} to synthesize training tasks and inputs, and \emph{directly} generate ground-truth task outputs based on its created tasks. 
However, these methods are not suitable for generating high-quality annotations for training agents. 
For example, given a web browsing task in Mind2Web, GPT-4 only achieves around 20\% step success rate~\citep{liu2023agentbench} when completing the task. Relying on such methods to generate complex interactive task annotations may degrade the annotation quality.

Instead of creating annotations with LLMs from scratch, we exploit LLMs as a ``style transfer'' tool to convert ground-truth reasoning steps in existing benchmarks into the expected format in \methodname{} formulations. There are a considerable number of the datasets annotated with either human-written solutions or structured action sequences\footnote{More available resources for future extension and the discussion about the scalability of our annotation conversion methods are described in App. \ref{app:possible_annot}.}. For example, PRM800K~\citep{lightman2023let} is a math dataset containing the natural language solutions interleaved with formulas; StrategyQA~\citep{geva-etal-2021-aristotle} are a QA dataset with decomposed questions, supporting facts, and relevant Wikipedia paragraph indices; Mind2Web includes ground-truth action sequences, etc. They provide LLMs with fundamental information that sufficiently contributes to the annotation conversion.

Next, we introduce 1) how we prompt LLMs to obtain the subgoal and action supervisions for training modules; 2) how to organize the subgoals and actions into the conversational forms aligning with \methodname{} formulations; 3) how we train the modules with the final annotations.

\subsection{Annotation Conversion Prompts}
\label{sec:annotations_conversion_prompts}
To help LLMs better follow the annotation conversion instructions, we add 4-/5-shot examples in conversion prompts (see App.~\ref{app:conversion-prompts} for prompt details). We discuss the important elements in these in-context examples. The notations of the converted annotations have hat over letters.


\paragraph{Action Interfaces.}
Action interfaces define the available actions that LLMs could ground to.  Table~\ref{table:appendix-tools} shows a comprehensive list of action definitions and their implementations.

\paragraph{Ground-Truth Intermediate Reasoning Steps.}
We provide LLMs with ground-truth intermediate reasoning steps in existing benchmarks. With these as references, LLMs are able to summarize high-level subgoals and synthesize corresponding actions according to the given action interfaces. 

\paragraph{Subgoals and Corresponding Actions.}
When converting ground-truth reasoning steps into our expected annotations, we provide LLMs with examples about how to distill the high-level subgoals from the reasoning steps and map them into corresponding actions. In the in-context examples, we manually decompose a complex task into high-level subgoals according to the context of ground-truth reasoning steps. Under each high-level subgoal, we write down multiple corresponding actions that help to accomplish the subgoal (shown in App.~\ref{app:conversion-prompts}). Given the exemplar subgoals and actions in the prompt, LLMs would emulate to generate subgoals and their paired actions when performing the conversion for new tasks.

As the executed results of prior subgoals might be useful in future action implementation, we interlink the grounded actions in the in-context examples to allow context-dependent execution. One example of the interlinked actions is \texttt{R1 = KnowledgeQuery(Zombies); R2 = ParagraphRetrieve(\hlcyan{R1}, Query: What color skin are zombies typically depicted with?)}. The agent could first find the zombie knowledge page (\texttt{\hlcyan{R1}}). Written in interlinked style, the \texttt{ParagraphRetrieve} action is able to receive the knowledge about zombies \texttt{\hlcyan{R1}} as the context, and performs query-based retrieval. 

\paragraph{Intermediate Executed Results of Subgoals.}
The intermediate executed results $\hat{E}$ play an important role in increasing \methodname{}'s adaptability to environmental changes. Some datasets (e.g., GSM8K) offer execution results in their reasoning steps, i.e., the computation results of formulas. For the datasets without any execution results, their reasoning steps actually contain the relevant clues for the execution results. We take an example in StrategyQA. Though the answer of the annotated decomposed question ``\texttt{What color skin are zombies typically depicted with?}'' is not directly provided, the annotation contains a related fact ``\texttt{Zombies are often depicted as green in pallor.}'' that mentions the answer ``\texttt{green}''. Thus, for each in-context example, we concatenate the relevant documents as well as our manually captured execution results in the conversion prompts. When converting new samples into \methodname{} annotations, LLMs would automatically extract the executed results from the given documents. 

After prompting LLMs with the conversion prompts, we can acquire the key elements in training annotations, including subgoals $\hat{S}$, their corresponding actions $\hat{A}$ and execution results $\hat{E}$. 

\subsection{Organizing Conversational Annotations}
\label{sec:annotations_transfer}
Finally, to build the interaction between planning and grounding modules, we organize the annotations into conversational format.

\paragraph{Conversational Planning Module Annotation.} 
As shown in App.~\ref{app:annotation_org}'s Fig.~\ref{fig:annotations-final-planning}, we first play a user role to provide the task $\hat{T}$ in the user prompt. For \methodnameo{}, all the subgoals $\hat{S}$ are the planning module's final outputs. 
\methodnamei{} requires multi-turn conversational style. From Fig.~\ref{fig:annotations-final-planning}, the planning module appends the first ground-truth subgoal $\hat{s}_1$ with index ``\texttt{Subgoal 1}'' as the first response supervision. We then put Subgoal 1's executed result $\hat{e}_1$ with prefix ``\texttt{The executed result for Subgoal 1 is}'' as the second user input. For the remaining turns, we act as the user, provide the execution results $\hat{e}_{t-1}$ of the last subgoal $\hat{s}_{t-1}$ to the planning module, and ask if the planning should be stopped. The response supervisions cover whether the planning should be terminated; if no, the response should contain a new gold subgoal $\hat{s}_t$.

\paragraph{Conversational Grounding Module Annotation.} 
As shown in App.~\ref{app:annotation_org}'s Fig.~\ref{fig:annotations-final-grounding}, we also first provide the task $\hat{T}$ and action interfaces $\hat{I}$ to the grounding module in the first user turn. For \methodnameo{}, we feed all the converted subgoal annotations $\hat{S}$ in the user prompt. All the action annotations $\hat{A}$ would be the user prompt response. 
For \methodnamei{}, we input the current gold subgoal $\hat{s}_t$, with prefix ``\texttt{Subgoal to be grounded:}’’. Its response would be $\hat{s}_t$’s corresponding actions $\hat{A}_t$.

\subsection{Training with Converted Annotations}
\label{sec:annotations_training}
As \methodname{} annotations are conversational, we formulate them as $\{x_1, y_1, ... , x_i, y_i, ... , x_n, y_n\}$, where $x_i$ is $i$-th user prompt and $y_i$ indicates its ground-truth responses. Following \citet{wang2023far}, during training, we feed each entire multi-turn annotation into a decoder-only model while merely calculating the decoding loss on the tokens of ground-truth responses $Y = \{y_1, ... , y_i, ..., y_n\}$. We apply binary masking on the user prompt tokens to prevent computing loss on them. The final loss function is 
$L = - \sum_j \log p_{\pi}(t_{j} \mid t_{< j} ) \times \mathbf{1}({t_j\in Y})$
where $t_j$ denotes $j$-th input token and $\mathbf{1}(\cdot)$ is a Boolean indicator function.

\section{Experiments}
\label{sec:exp}

We begin with the details of our experimental setup, including annotation conversion, module training, and the tools used in the execution module. We then evaluate \methodname{} by: 1) comparing \methodname{} with existing open-source LLM agents and GPT-based agents, 2) contrasting \methodname{} against other agent training methods, 3) manifesting \methodname{} generalizability on two unseen tasks involving new environments and actions, and 4) assessing \methodname{} annotation quality.

\subsection{Experimental Setup}
\label{sec:exp-setups}
\paragraph{Data Collection.}
Utilizing the conversion prompts discussed in \S\ref{sec:annotations_conversion_prompts}, we employ GPT-4 \citep{achiam2023gpt} versions on 8/13/2023 and 9/13/2023, and GPT-4V \citep{2023GPT4VisionSC} version on 1/24/2023 to perform annotation conversion on the ground-truth reasoning steps in existing benchmarks. App.~\ref{app:stats} provides the data sources used for annotation conversion. These include the datasets of math, QA, web and multimodal tasks. To help \methodname{} be aware of the visual inputs in multimodal tasks, we append a detailed image caption generated by LLAMA-1.5-7B \cite{liu2023improved} to the task description in train and test sets. After filtering out the ones that contain mismatched parentheses, invalid execution outputs or excessively lengthy outputs, we obtain 55,382 and 55,499 annotations for training the planning and grounding modules, respectively.

\paragraph{Training and Action Interfaces.}
We adopt LLAMA-2-7B and LLAMA-2-13B~\cite{touvron2023llama} as the base models for both the planning and grounding modules. Details regarding the training process can be found in App.~\ref{app:training}. For solving interactive tasks, we integrate commonly used actions for each task into the pre-defined action interfaces. Details of supported executable actions are included in App.~\ref{app:tools}.

\begin{table*}[ht]
    \centering
    \vspace{-2.5em}
    \begin{subtable}[ht]{0.26\textwidth}
    \centering
    \scalebox{0.61}{
    \begin{tabular}{cc}\toprule
    \multirow{2}{*}{\textbf{Agents}} & \textbf{Web Task} \\
    \cmidrule(lr){2-2}
    &\hlpink{Mind2Web} \\ \midrule
    \rowcolor[gray]{0.95} \multicolumn{2}{c}{\texttt{GPT/API-based  Agents}}  \\ \midrule
    GPT-3.5-turbo & 15.7 \\
    GPT-4 & 22.6 \\
    \midrule
    \rowcolor[gray]{0.95} \multicolumn{2}{c}{\texttt{Open-source  Agents}}  \\ \midrule
    Baichuan-13B-chat & 2.3 \\
    WizardLM-30B & 3.1 \\
    Koala-13B & 6.0 \\
    AgentLM-70B & 13.5$^\bigstar$ \\
    \midrule
    \methodnamei{}$_{\mathrm{Web}}$ & 27.6$^\bigstar$ \\
    \methodnamei{}$_{\mathrm{Web}}$-13B & \textbf{31.3}$^\bigstar$ \\
    \bottomrule
    \end{tabular}
    }
    \caption{\small \textbf{Web} performance in step-wise success rate.}
    \label{table:exp-web-agent}
    \end{subtable}
    \hspace{12pt}
    \begin{subtable}[ht]{0.3\textwidth}
    \centering
    \scalebox{0.61}{
    \begin{tabular}{ccc}\toprule
    \multirow{2}{*}{\textbf{Agents}} &\multicolumn{2}{c}{\textbf{Math Tasks}} \\\cmidrule{2-3}
    &\hlpink{GSM8K} &\hlcyan{SVAMP} \\\midrule
    \rowcolor[gray]{0.95} \multicolumn{3}{c}{\texttt{Open-source Agents}}  \\ \midrule
    AgentLM-13B & 32.4 & - \\
    Code-Llama (PoT)-13B &36.1 &60.0 \\
    Platypus-30B &37.8 &51.7 \\
    ReWOO-open & $\approx$38 &- \\
    Orca-Platypus-13B &38.4$^\bigstar$ &56.9 \\
    Alpaca-7B & $\approx$39 &- \\
    Galactica-30B &41.7 &41.6 \\
    \midrule
    \methodnameo{}$_{\mathrm{Math}}$ &50.5$^\bigstar$ &65.5 \\
    \methodnamei{}$_{\mathrm{Math}}$ &47.1$^\bigstar$ &63.6 \\
    \methodnameo{}$_{\mathrm{Math}}$-13B &\textbf{55.4}$^\bigstar$ &\textbf{69.3} \\
    \bottomrule
    \end{tabular}
    }
    \caption{\small \textbf{Math} performance in accuracy. }
    \label{table:exp-maths}
    \end{subtable}
    \hspace{12pt}
    \begin{subtable}[ht]{0.35\textwidth}
    \centering
    \scalebox{0.56}{
    \begin{tabular}{ccc}\toprule
    \multirow{2}{*}{\textbf{Agents}} &\multicolumn{2}{c}{\textbf{Multimodal Tasks}} \\\cmidrule{2-3}
    &\hlpink{A-OKVQA} &\hlcyan{ScienceQA (IMG)} \\ \midrule
    \rowcolor[gray]{0.95} \multicolumn{3}{c}{\texttt{GPT/API-based Agents}}  \\ \midrule
    GPT-3 + GT Caption & 45.4 & - \\ \midrule
    \rowcolor[gray]{0.95} \multicolumn{3}{c}{\texttt{Open-source Agents}}  \\ \midrule
    ClipCap (VL) & 44.0$^\bigstar$ & - \\
    KRISP (VL) & 51.9$^\bigstar$ & - \\
    GPV-2 (VL) & 60.3$^\bigstar$ & - \\
    MiniGPT-4-13B (VL) & 67.2 & 42.8 \\
    LLAVA-1.5-7B (VL) & - & 57.6 \\
    \midrule
    \methodnameo{}$_{\mathrm{MM}}$ &70.1$^\bigstar$ & 56.9 \\
    \methodnamei{}$_{\mathrm{MM}}$ &71.3$^\bigstar$ & \textbf{58.4} \\
    \methodnamei{}$_{\mathrm{MM}}$-13B &\textbf{72.4}$^\bigstar$ & 58.2 \\
    \bottomrule
    \end{tabular}
    }
    \caption{\small \textbf{Multimodal} performance in accuracy.}
    \label{table:exp-multimodal}
    \end{subtable}
    \hfill
    \vspace{6pt}
    
    \begin{subtable}[ht]{0.65\textwidth}
    \vspace{-0.3em}
    \scalebox{0.61}{
    \begin{tabular}{ccccc}\toprule
    \multirow{2}{*}{\textbf{Agents}} &\multirow{2}{*}{\textbf{Agent Model}} &\multirow{2}{*}{\textbf{QA Tool}} &\multicolumn{2}{c}{\textbf{QA Tasks}} \\\cmidrule{4-5}
    & & &\hlpink{StrategyQA} & \hlcyan{HotpotQA} (LLM Acc. / EM) \\\midrule
    
    \rowcolor[gray]{0.95} \multicolumn{5}{c}{\texttt{GPT/API-based Agents}}  \\ \midrule
    GPT-3.5-CoT & GPT-3.5-turbo & GPT-3.5-turbo &56.0 & 37.8 / 22.4 \\
    ReAcT &GPT-3.5-turbo &GPT-3.5-turbo &64.6 & 40.8 / 32.4 \\
    ReWOO &GPT-3.5-turbo &GPT-3.5-turbo &66.6 & 42.4 / 30.4 \\ \midrule
    
    \rowcolor[gray]{0.95} \multicolumn{5}{c}{\texttt{Open-source Agents}}  \\ \midrule
    ReWOO-open & LLAMA-7B &GPT-3.5-turbo & $\approx$56 & $\approx$37 / -$^\bigstar$ \\
    AgentLM & LLAMA-2-7B &LLAMA-2-7B & - & - / 22.3 \\
    FiReAct & LLAMA-2-7B &LLAMA-2-7B & - & - / 26.2$^\bigstar$ \\
    FiReAct & CodeLLAMA-34B &CodeLLAMA-34B & - & - / 27.8$^\bigstar$ \\
    \midrule
    \methodnameo{}$_{\mathrm{QA}}$ &LLAMA-2-7B &GPT-3.5-turbo &60.6$^\bigstar$ & 39.2 / 24.9 \\
    \methodnamei{}$_{\mathrm{QA}}$ &LLAMA-2-7B &LLAMA-2-7B & 58.3$^\bigstar$ & 37.3 / 23.5 \\
    \methodnamei{}$_{\mathrm{QA}}$ &LLAMA-2-7B &GPT-3.5-turbo &65.7$^\bigstar$ & 45.9 / 29.4 \\
    \methodnamei{}$_{\mathrm{QA}}$ &LLAMA-2-7B &GPT-4 &72.4$^\bigstar$ & 56.8 / 36.0 \\
    \methodnamei{}$_{\mathrm{QA}}$ &LLAMA-2-13B &GPT-3.5-turbo &65.3$^\bigstar$ & 50.2 / 31.4 \\
    \methodnamei{}$_{\mathrm{QA}}$ &LLAMA-2-13B &GPT-4 &\textbf{76.7}$^\bigstar$ & \textbf{57.4} / \textbf{36.3} \\
    \bottomrule
    \end{tabular}
    }
    \caption{\small \textbf{QA} performance. The evaluation metric for StrategyQA and HotpotQA is accuracy, and LLM accuracy / Exact Match (EM), respectively.}
    \label{table:exp-complex-qa}
    \end{subtable}
    \hspace{6pt}
    \begin{subtable}[ht]{0.32\textwidth}
    \centering
    \scalebox{0.63}{
    \begin{tabular}{ccc}\toprule
    \multirow{2}{*}{\textbf{Agents}} &\multicolumn{2}{c}{\textbf{Unseen Tasks}} \\
    \cmidrule(lr){2-3}
    &\hlcyan{WebShop} &\hlcyan{InterCode$_{\mathrm{SQL}}$}\\\midrule
    Baichuan-13B-chat & 5.7 & - \\
    Koala-13B &6.0 & - \\
    WizardLM-30B &10.6 & - \\
    Vicuna-v1.1-13B &12.6 & - \\
    ChatGLM2 &19.4 & - \\ 
    Vicuna-v1.3-33B &23.9 & 6.7 \\
    Vicuna-v1.5-13B &41.7 & 4.8 \\ 
    OpenChat-v3.2-13B &46.9 & - \\
    Claude-instant &49.7 & - \\ \midrule
    \methodnamei{}$_{\mathrm{Web}}$-13B & 46.2 & 4.2 \\
    \methodnamei{}$_{\mathrm{Math}}$-13B & 45.7 & 5.8 \\
    \methodnamei{}$_{\mathrm{QA}}$-13B & 47.3 & 3.5 \\
    \methodnamei{}$_{\mathrm{MM}}$-13B & 43.8 & 4.0 \\ \midrule
    \methodnamei{}$_{\mathrm{All}}$-13B &\textbf{50.3} & \textbf{7.3} \\ \bottomrule
    \end{tabular}
    }
    \caption{\small Unseen tasks, \textbf{WebShop} and \textbf{InterCode}$_{\mathrm{SQL}}$. The metric is average reward and success rate, respectively.}
    \label{table:exp-unseen}
    \end{subtable}
    \hfill
    \caption{Overall performance of language agents on diverse complex interactive tasks. The tasks highlighted in \hlpink{red} and \hlcyan{blue} are the held-in/held-out datasets for the trained task types. $^\bigstar$ indicates that the results are obtained after fine-tuning on the task's training set. We adopt multiple-choice setting for A-OKVQA. IMG denotes the subset with image inputs in ScienceQA test set. GPT-3 in Tab. \ref{table:exp-multimodal} indicates \texttt{text-davinci-002}.
    }
    \vspace{-9pt}
    \label{table:exp-overall}
\end{table*}



\subsection{Training Task Performance}
\label{sec:exp-training-task-perf}
We evaluate \methodname{} across an array of complex interactive tasks - QA, web, math and multimodal tasks. The evaluation mainly follows the settings established by AgentBench~\citep{liu2023agentbench} and ReWOO~\citep{xu2023rewoo} (see App.~\ref{app:eval}). For each task type, excluding web tasks, we include a held-out dataset to assess the model's generalizability across the domains within the same task category. The performance is displayed in Tab.~\ref{table:exp-overall}. 
Note that in Tab.~\ref{table:exp-overall}, task-specific agents \methodnamei{}$_{\mathrm{X}}$ are trained using task-specific data belonging to task type $\mathrm{X}$ (e.g., $\mathrm{Web}$, $\mathrm{Math}$, $\mathrm{QA}$, $\mathrm{MM}$). \methodnamei{}$_{\mathrm{All}}$ represents the agent after unified training with the combination of four task type annotations.
More evaluation details are shown in App.~\ref{app:eval}.

\paragraph{\methodname{} vs. Open-Source Agents.}
Overall, we find that \methodname{} consistently outperforms various open-source agents across the seven datasets. Though the base models of some compared agents are 2-10$\times$ larger than \methodname{}, \methodname{} significantly excels in performance. Specifically, 7B \methodnamei{} achieves 24.5\% and 14.1\% step success rate improvements over WizardLM-30B and AgentLM-70B on Mind2Web. 

The effectiveness of \methodname{} is particularly manifested on the held-out datasets which are unused in \methodname{} training data. Our observations reveal that \methodname{} outperforms many baseline open-source agents across all held-out datasets. Notably, even though Orca-Platypus-13B~\citep{hunterlee2023orcaplaty1} has been trained on a math corpus that includes GSM8K, its performance still 8.6\% lower than \methodnameo{} on SVAMP~\citep{patel-etal-2021-nlp}. Moreover, despite ReWOO and FiReAct being fine-tuned with in-domain HotpotQA annotations, \methodnamei{}, without any fine-tuning on HotpotQA, still presents an impressive improvement. A similar trend can be observed on ScienceQA. We compare AutoAct-7B~\citep{qiao2024autoact}, an open-source agent specifically trained on ScienceQA. \methodname{} achieves 67.3\% on the entire test set of ScienceQA, while AutoAct-7B's performance is only 53.3\%.

\paragraph{\methodname{} vs. GPT-based Agents.}
Although \methodname{} is built on LLAMA-2-7B/13B, \methodnamei{} delivers superior performance by 5.0-8.7\% over GPT-4 on Mind2Web. We also notice a 3.5-7.8\% increase in LLM accuracy over the GPT-based ReWOO on the HotpotQA dataset when employing GPT-3.5-turbo as the implementation of the QA tool to ensure fair comparisons. 



\subsection{Agent Formulation Comparison}
We train models using the same base model and data, but with different training methods -
\textbf{Chain-of-Thoughts (CoT) Training}: For a given task $T$, the agent learns to produce both the chain-of-thoughts solution and the final answer directly;
\textbf{Integrated Agent Training}: For a given task $T$, the agent learns to generate all the \textit{subgoals} and \textit{actions} using the same model. The execution modules remains the same. This training paradigm is adopted in ReWOO-open, FiReAct and AgentLM.


From Tab.~\ref{table:exp-formulations}, both \methodnamei{} and \methodnameo{} outperform CoT Training\footnote{We do not implement CoT training on web tasks, as updates to the environment (e.g., changes to HTML) are necessary intermediaries for planning subsequent actions.}. They also exceed the integrated formulation based on a single module to operate planning and grounding. 
It highlights the importance of disentangling subgoal planning and action grounding skills during the agent training.


 

\subsection{\methodname{} Generalizability Evaluation}
\label{sec:exp-generalizability}
 
Since \methodname{} employs a unified format to represent complex interactive tasks, we envision that after trained with the combined annotations across the four training task types, i.e., unified training, when faced with an unseen task type, \methodname{} may adapt to it more effectively.



\begin{table}[t]
    \centering
    \scalebox{0.68}{
    \begin{tabular}{lcc}\toprule
    \multirow{2}{*}{\textbf{Training Data}} &\multicolumn{2}{c}{\textbf{QA}} \\ \cmidrule{2-3}
    &StrategyQA &HotpotQA \\ \midrule
    \rowcolor[gray]{0.95} \multicolumn{3}{l}{\texttt{Downstream Perf. of Training Different Data w/ LLAMA}}  \\ \midrule
    \makecell[l]{ReWOO-open Data} & $\approx$57 & $\approx$37 \\
    \methodnamei{}$_{\mathrm{QA}}$ Data &\textbf{58.3} & \textbf{38.1}\\ \midrule
    \rowcolor[gray]{0.95} \multicolumn{3}{l}{\texttt{Perf. Using High- \& Low-Level Subgoal Annots. w/ LLAMA-2}}  \\ \midrule
    \methodnamei{}$_{\mathrm{QA}}$ w/ Low-Level Subgoals & 63.3 & 44.3 \\
    \methodnamei{}$_{\mathrm{QA}}$ Data &\textbf{65.7} & \textbf{45.9}\\
    \bottomrule
    \end{tabular}
    }
    \caption{Comparison between the 7B-sized agents trained with different annotations.}
    \vspace{-9pt}
    \label{table:exp-annotation-analysis}
\end{table}

To examine the generalization ability of \methodname{}, we evaluate it on the unseen tasks, WebShop~\citep{yao2022webshop} and InterCode$_{\mathrm{SQL}}$. 
WebShop resembles a text game\footnote{WebShop utilizes four actions in its training annotations: \texttt{Search}, \texttt{FeatureRetrieve}, \texttt{Pick}, and \texttt{Click}. The argument of \texttt{Click} is a shopping item, differing from that of \texttt{Click} in Mind2Web which includes an HTML element description.}, with its shopping environment and action space considerably differing from those covered in the training annotations of \methodname{}.  InterCode$_{\mathrm{SQL}}$~\citep{yang2023intercode} is an interactive code generation task that requires the agent to generate SQL code based on the external databases and involves unseen SQL commands. To make \methodname{} adapt to these unseen tasks, we supplement 2-/3-shot examples in the module inputs, enabling them to learn how to generate subgoals and ground to new sets of available actions (see App.~\ref{app:webshop}).

\begin{table*}[t]
    \vspace{-2em}
    \begin{subtable}[t]{0.18\textwidth}
    \centering
    \scalebox{0.54}{
    \begin{tabular}{lc}\toprule
    \multirow{2}{*}{\textbf{Agents}} & \textbf{Web Task} \\ \cmidrule{2-2}
    &Mind2Web \\ \midrule
    Integrated Training & 25.3 \\
    \midrule
    \methodnamei{}$_{\mathrm{Web}}$ & \textbf{27.6} \\
    \bottomrule
    \end{tabular}
    }
    \caption{Web task.}
    \label{table:exp-formulations-web-agent}
    \end{subtable}
    \hspace{6pt}
    \begin{subtable}[t]{0.22\textwidth}
    \centering
    \scalebox{0.54}{
    \begin{tabular}{lcc}\toprule
    \multirow{2}{*}{\textbf{Agents}} &\multicolumn{2}{c}{\textbf{Math Tasks}} \\\cmidrule{2-3}
    &GSM8K &SVAMP \\\midrule
    CoT Training & 40.4 & 52.2 \\
    Integrated Training & 45.5 & 61.7 \\
    \midrule
    \methodnameo{}$_{\mathrm{Math}}$ &\textbf{50.5} & \textbf{65.5} \\
    \methodnamei{}$_{\mathrm{Math}}$ &47.1 & 63.6 \\
    \bottomrule
    \end{tabular}
    }
    \caption{Math tasks.}
    \label{table:exp-formulations-math}
    \end{subtable}
    \hspace{6pt}
    \begin{subtable}[t]{0.25\textwidth}
    \centering
    \scalebox{0.54}{
    \begin{tabular}{lcc}\toprule
    \multirow{2}{*}{\textbf{Agents}} &\multicolumn{2}{c}{\textbf{QA Tasks}} \\ \cmidrule{2-3}
    &StrategyQA &HotpotQA \\ \midrule
    CoT Training & 58.3 & 22.1 \\
    Integrated Training & 62.3 & 39.6 \\
    \midrule
    \methodnameo{}$_{\mathrm{QA}}$ &60.6 & 39.2\\
    \methodnamei{}$_{\mathrm{QA}}$ &\textbf{65.7} & \textbf{45.9}\\
    \bottomrule
    \end{tabular}
    }
    \caption{QA tasks.}
    \label{table:exp-formulations-complex-qa}
    \end{subtable}
    \hspace{6pt}
    \begin{subtable}[t]{0.25\textwidth}
    \centering
    \scalebox{0.54}{
    \begin{tabular}{lcc}\toprule
    \multirow{2}{*}{\textbf{Agents}} &\multicolumn{2}{c}{\textbf{Multimodal Tasks}} \\ \cmidrule{2-3}
    &A-OKVQA &ScienceQA \\ \midrule
    CoT Training & 68.3 & 57.3 \\
    Integrated Training & 70.9 & 57.6 \\
    \midrule
    \methodnameo{}$_{\mathrm{MM}}$ & 70.1 & 56.9 \\
    \methodnamei{}$_{\mathrm{MM}}$ & \textbf{71.3} & \textbf{58.4} \\
    \bottomrule
    \end{tabular}
    }
    \caption{Multimodal tasks.}
    \label{table:exp-formulations-multimodal}
    \end{subtable}
    \caption{Comparison among different formulations of training language agents. The metric for HotpotQA is LLM accuracy (\%). All the experiments are based on LLAMA-2-7B.}
    \vspace{-9pt}
    \label{table:exp-formulations}
\end{table*}


As outlined in Tab.~\ref{table:exp-unseen}, \methodnamei{} achieves a 3-7 higher average reward than domain-specific agents on WebShop. It also significantly outperforms larger agents such as WizardLM-30B and Vicuna-v1.3-33B \citep{vicuna2023}.
We observe the similar trend on InterCode$_{\mathrm{SQL}}$. It suggests that unified training enables agents to enhance the generalization on unseen tasks and novel actions. We provide more unified training results in App.~\ref{app:unified} to show unified training also boosts the performance on most training task types.



\subsection{Further Analysis on Training Annotations}
We aim to address three questions pertinent to quality and format decisions. Q1: How good are our converted training annotations? Q2: Would adopting low-level subgoals be more effective than using high-level subgoals? Q3: Would incorporating \methodname{} annotation with general instruction-tuning data achieve comparable performance on instruction-following benchmarks?


\paragraph{Assessment of Annotation Quality.}
We assess the quality of our annotations by training models with them and evaluating the agents' performance. We compare them with ReWOO-open data, constructed based on HotpotQA and TriviaQA~\citep{joshi-etal-2017-triviaqa} using the Self-Instruct method. For fair comparison, we train the base model of ReWOO-open, LLAMA-7B, using \methodname{} annotations. We also adopt the integrated training formulation and sample 2,000 training data to keep the same training settings as ReWOO-open. Given that ReWOO-open data exclusively relies on QA benchmarks, we primarily focus on QA task evaluation.
Shown in Tab.~\ref{table:exp-annotation-analysis}, \methodname{} data yields an improvement when compared to ReWOO-open data on StrategyQA and HotpotQA. Note that even if the ReWOO-open data are based on HotpotQA, it still underperforms \methodname{} on HotpotQA. 

\paragraph{Low-Level Subgoal vs. High-Level Subgoal.}
As described in \S\ref{sec:formulations}, we ask LLMs to generate high-level subgoals corresponding to one or many low-level actions. An alternative annotation could be one where each subgoal corresponds solely to one low-level action, i.e., the subgoal is ``low-level''. We direct LLMs to create low-level subgoals by modifying the annotation conversion prompt to fit the format where a subgoal is strictly linked to one action. Tab. \ref{table:exp-annotation-analysis} reveals a drop after replacing high-level subgoals with low-level ones on both QA datasets. This result hence reaffirms the appropriateness of our initial subgoal design.

\paragraph{Effect on General Instruction-Following Benchmark.}
As \methodname{} adopts the conversational style discussed in Sec. \ref{sec:annotations}, any tasks that can be formalized as conversation tasks could be accommodated. Therefore, \methodname{} supports all the instruction-tuning tasks and complex interactive tasks. We conduct an experiment to show the effectiveness of \methodname{} in instruction tuning tasks. By integrating the general instruction-tuning training data, Alpaca, with the unified training annotations of \methodname{}, we created an expanded training dataset. To guide \methodname{} towards generating straightforward responses, we appended the phrase ``\texttt{Respond to me directly}'' to each input instance from Alpaca, instructing the model to avoid elaborating reasoning processes for the general instruction-tuning tasks. Tested on the Super-NaturalInstruction benchmark~\cite{wang-etal-2022-super}, Lumos achieved a 39.3 ROUGE-L score, which is close to the LLAMA-2-7B trained on Alpaca alone (39.8), which suggesting the possibility of developing agent reasoning ability and general instruction following ability simultaneously.

\section{Related Work}
\paragraph{LLM Agents.}
Language agents have shown potential in solving diverse complex interactive tasks. ReAct~\citep{yao2022react} introduced a prompting method that shaped LLMs as language agents and grounded them in external environments. Subsequently, several methods~\citep{shen2023hugginggpt,lu2023chameleon,xu2023rewoo,Lin2023SwiftSageAG,liu2023bolaa} aimed at improving agent performance and increasing their applicability in diverse scenarios. These agents mainly rely on closed-source LLMs, lacking of the consideration of affordability, reproducibility and transparency issues on complex interactive tasks. 

\paragraph{Improving Small Models for Building Agents.}
Recent works have utilized larger models to generate training data for fine-tuning smaller models~\citep{bosselut-etal-2019-comet,west-etal-2022-symbolic,wang-etal-2023-self-instruct,hsieh-etal-2023-distilling,brahman2023plasma} to enable them to follow instructions and perform chain-of-thoughts reasoning. We also observe contemporaneous efforts ReWOO-open \citep{xu2023rewoo}, FireAct \cite{chen2023fireact}, AgentLM \citep{zeng2023agenttuning}, and AutoAct \cite{qiao2024autoact}, focusing on training agents on smaller LLMs. Unlike FireAct and AutoAct, our work delves into a more in-depth analysis, aiming to discover a unified task representation that enables agents to generalize across unseen interactive tasks effectively. In contrast to ReWOO-open and AgentLM, we extend to examining proper training formulations and studying multiple strategies for creating large-scale, high-quality datasets for agent training. We demonstrate \methodname{} superior performance in \S \ref{sec:exp}. 



\section{Conclusion}
We introduce \methodname{}, an open-source, generalizable language agent training framework. We propose two formulations, \methodnamei{} and \methodnameo{}, which promote collaboration among agent modules to solve complex tasks. For module training data, we use LLMs to transform reasoning steps in existing benchmarks into a unified format applicable within \methodname{} framework. \methodname{} outperforms a variety of open-source agents across the 9 datasets. It performs even better than GPT agents on QA and web tasks. \methodname{} also exceeds potential agent training formulations and exhibits superior generalization on two unseen interactive tasks.

\section*{Limitations}
\paragraph{Covered Training Task Types.}
Currently, \methodname{} is trained using annotations for four specific types of complex interactive tasks, which may still limit its generalization capabilities for novel tasks. To address this, we aim to enrich the training data for \methodname{} by incorporating a wider variety of task types. As outlined in \S \ref{sec:annotations} and App. \ref{app:possible_annot}, a substantial array of benchmarks already exists, providing ground-truth reasoning steps that could serve as a foundation for expanding \methodname{}'s annotations. By broadening the scope of annotations, we not only enhance the language agents but also offer a valuable resource for practitioners looking to develop their own models.

\paragraph{Backtracking and Replanning Ability.}
In situations where language agents encounter invalid execution outcomes or navigate erroneous solution pathways, it is crucial for them to possess the capacity for self-diagnosis and replanning their reasoning processing. The current \methodname{} lacks these sophisticated self-corrective features. Future versions should be designed with advanced mechanisms that enable the agents to recognize and rectify their planning errors.

\paragraph{Open-Source Tool Replacement.} For part of our QA experiments, we employ GPT models to address decomposed sub-questions. It is designed for fair comparison with the agents that also use GPT models as QA tools, as elaborated in \S \ref{sec:exp-training-task-perf}. Our future strategy involves transitioning to fully open-source QA frameworks that leverage models such as LLAMA-2-70B, aiming to establish a completely open-source framework.

\section*{Acknowledgement}
This work was funded in part by DARPA MCS program through NIWC Pacific (N66001-19-2- 4031), DARPA ECOLE (\#HR00112390060), and the Allen Institute for
AI. We thank Mosaic team members, and the anonymous reviewers
for the helpful discussions.


\bibliography{custom}

\clearpage
\appendix
\section*{Appendix}
\section{Illustration of Annotation Organization}
\label{app:annotation_org}
\begin{figure*}[t]
\centering
\begin{subfigure}{\textwidth}
\centering
\scalebox{1}{
\includegraphics[width=\textwidth, trim=10 30 20 0, clip]{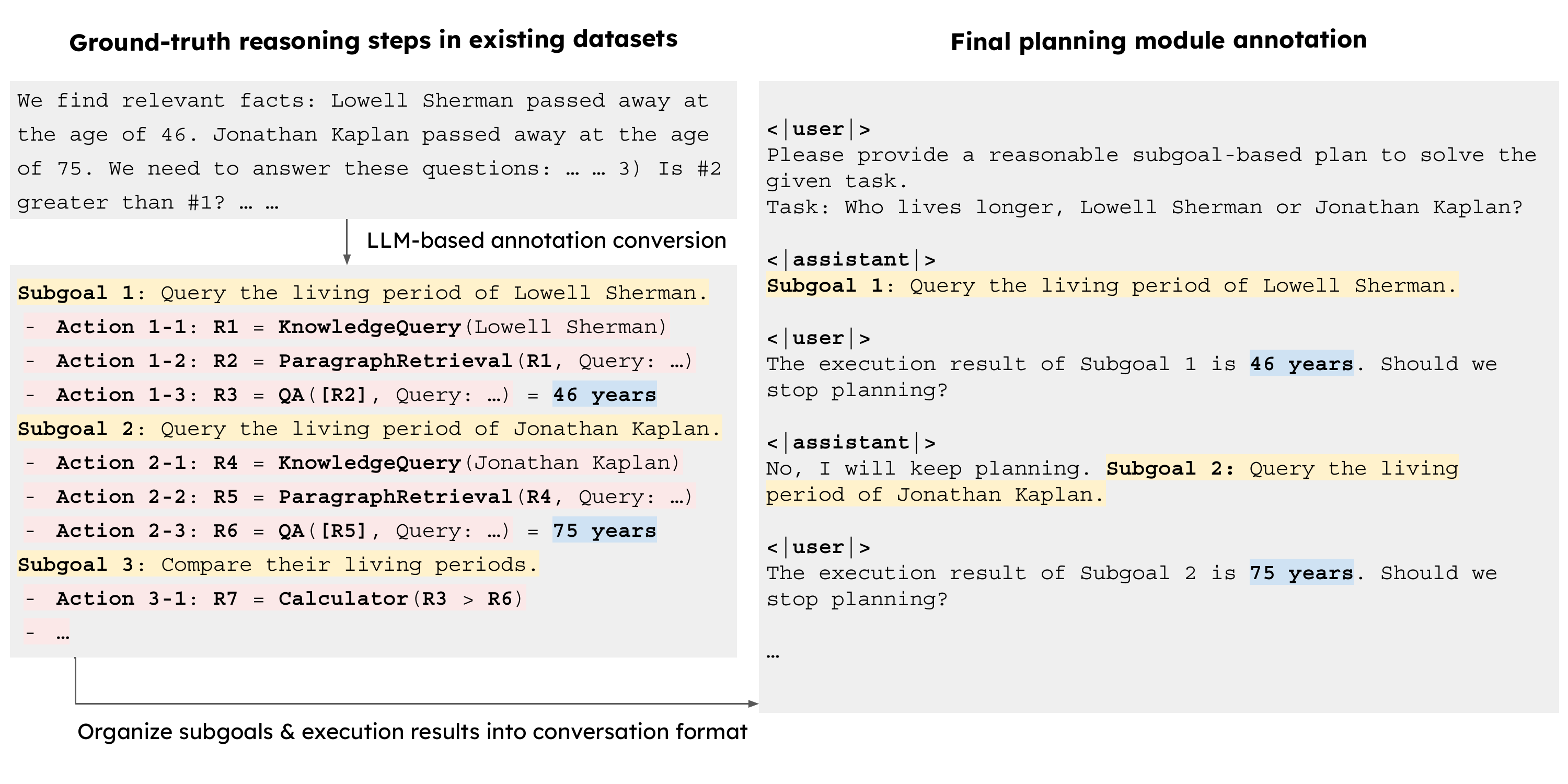}
}
\caption{Final planning module annotation organized from the converted subgoals \& execution results.}
\label{fig:annotations-final-planning}
\end{subfigure}
\hfill

\begin{subfigure}{\textwidth}
\centering
\scalebox{1}{
\includegraphics[width=\textwidth, trim=0 0 20 0, clip]{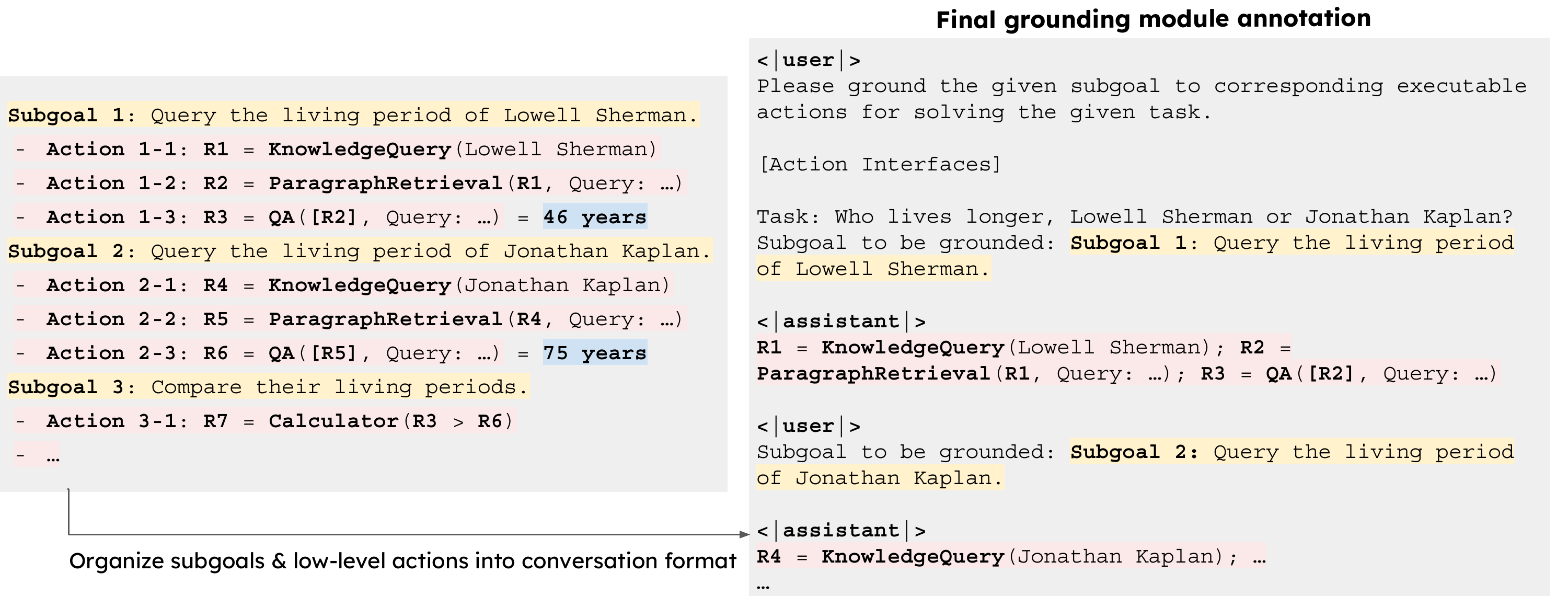}
}
\caption{Final grounding module annotation organized from the converted subgoals \& actions.}
\label{fig:annotations-final-grounding}
\end{subfigure}
\caption{Process of converting converted subgoals, actions, and executions into the final conversational training annotations for \methodnamei{} formulation.}
\vspace{-6pt}
\label{fig:annotations-final}
\end{figure*}

As discussed in \S\ref{sec:annotations_transfer}, we organize the ground-truth subgoals and actions converted by GPT-4 into the conversational format that boosts the interaction between modules. We show the final conversation format in Fig.~\ref{fig:annotations-final}.

\section{Statistics of Converted Training Annotations}
\label{app:stats}
As discussed in \S\ref{sec:exp-setups}, the data sources for constructing training annotations cover a broad range of complex interactive tasks. Tab.~\ref{table:appendix-stats} shows the benchmarks leveraged for annotation conversion, along with the task type information. 

To train agents like \methodnamei{}$_{\mathrm{Math}}$ mentioned in Tab.~\ref{table:exp-maths}, we need to leverage the annotations converted from 19778 data specific to math domain. For training a unified agent such as \methodnamei{}, we would use the annotations transformed from all the listed data as training set.

We calculate the average turn numbers in each task's converted training annotations. The average numbers are 4.75, 3.75, 8.25 and 3.92 for Math, QA, Web and Multimodal tasks, respectively.

\begin{table*}[h]
\centering
\scalebox{0.65}{
\begin{tabular}{cccccc}
\toprule
\textbf{Task Types}                  & \textbf{Datasets}   & \textbf{\# Source Data} & \textbf{\# Total} & \textbf{\# Final Converted for Planning} & \textbf{\# Final Converted for Grounding} \\ \midrule
\multirow{3}{*}{Math}      & PRM800K~\citep{lightman2023let}    & 10000   & \multirow{3}{*}{19778}   &   \multirow{3}{*}{19386}   &   \multirow{3}{*}{19471}       \\
& GSM8K~\citep{cobbe2021training}      & 7473       &      &       \\
& ASDiv~\citep{miao-etal-2020-diverse}      & 2305      &    &          \\ \midrule
\multirow{2}{*}{QA} & Musique~\cite{trivedi-etal-2022-musique}    & 17632   & \multirow{2}{*}{19409}   & \multirow{2}{*}{19048}   & \multirow{2}{*}{19080}          \\
& StrategyQA~\citep{geva-etal-2021-aristotle} & 1777   &   &             \\ \midrule
Web                   & Mind2Web~\citep{deng2023mind2web}   & 1009 & 1009 & 1007  & 1007  \\ \midrule
Multimodal                   & A-OKVQA~\citep{schwenk2022okvqa}   & 17056 & 17056 & 15941   & 15941    \\
\bottomrule
\end{tabular}
}
\caption{Statistics of data sources for conversion and the number of final successfully converted annotations for each task type.}
\label{table:appendix-stats}
\end{table*}

\section{Available Resources for \methodname{} Training Data Extension}
\label{app:possible_annot}
As discussed in \S\ref{sec:annotations}, we seek the existing datasets with ground-truth intermediate reasoning steps to synthesize \methodname{} training annotations from four complex interactive task categories, math, QA, web and multimodal tasks.

The methodology for converting annotations as described is not limited to the four task types previously mentioned. Any training datasets that include gold reasoning rationales are suitable for the construction of annotations using the \methodname{} method. We present an exhaustive list of datasets that span a variety of task types in Tab.~\ref{table:appendix-possible-annot}. For example, the AlfWorld dataset requires actions that have not been encountered in existing \methodname{} annotations, such as \texttt{open}, \texttt{take}, \texttt{move}, and others; the TravelPlanner dataset requires actions like \texttt{CitySearch}, \texttt{FlightSearch}, which are equally unseen in the existing training set as well. This approach could significantly enhance the scalability of \methodname{} annotations, thereby augmenting the method's capability to adapt to novel environments and acquire proficiency in executing new actions.

\begin{table}[h]
\centering
\scalebox{0.63}{
\begin{tabular}{ccc}
\toprule
\textbf{Datasets}                  & \textbf{Task Types}   & \textbf{\# Total} \\ \midrule
WebLINX~\citep{lù2024weblinx}                   &  Web Browsing & 2337 \\
TravelPlanner~\citep{xie2024travelplanner}                   &  Travel Planning  & 225 \\
SciQ~\citep{welbl-etal-2017-crowdsourcing}                   &  Question Answering  & 11679 \\
GrailQA~\citep{gu2021beyond}                   &  Knowledge Graph Reasoning  & 44337 \\
NL2Bash~\citep{lin2018nl2bash}                   &  Interactive Coding  & 8090 \\
AlfWorld~\citep{shridhar2020alfworld}                   &  Embodied Task  & 3553 \\
VCR~\citep{zellers2019recognition}                   &  Multimodal Reasoning  & 212923 \\
LILA~\citep{mishra-etal-2022-lila}                   &  Math  & 93670 \\
\bottomrule
\end{tabular}
}
\caption{Statistics of potential data sources for \methodname{} annotation extension.}
\label{table:appendix-possible-annot}
\end{table}



\section{Details of Training Modules}
\label{app:training}
We describe additional details about our training experiments. In all of our experiments, we implement training over two epochs with a learning rate of $2\times10^{-5}$ and a batch size 128.  We set the maximum sequence length to 1024. We also apply linear warmup for 3\% of the total training steps to adjust the learning rate. All the training experiments are implemented with 2 NVIDIA 80GB A100 GPUs or 4 NVIDIA 48GB A6000 GPUs.

\section{Efficiency Study on \methodname{} and Efficiency-Performance Trade-Off}
\label{app:efficiency}
We compute the inference time for \methodnameo{} and \methodnamei{} across 100 instances on GSM8K and HotpotQA, respectively. The experiments are run with 2 NVIDIA A6000 48GB GPUs with inference batch size 16. As depicted in Tab. \ref{table:appendix-efficiency}, we find that  \methodnameo{} is much more efficient than \methodnamei{} on both datasets.

\begin{table}[h]
\centering
\scalebox{0.7}{
\begin{tabular}{ccc}
\toprule
\textbf{Agents} & \textbf{GSM8K}                  & \textbf{HotpotQA}   \\ \midrule
\methodnameo{} &    102s       & 556s  \\
\methodnamei{} &    851s    & 1007s  \\
\bottomrule
\end{tabular}
}
\caption{The time cost of \methodnameo{} and \methodnamei{} when performing inference 100 instances from GSM8K and HotpotQA datasets.}
\label{table:appendix-efficiency}
\end{table}

\methodnameo{} completes its inference in a single round, whereas \methodnamei{} necessitates multiple rounds of inference until it autonomously concludes its planning. The iterative planning and grounding in \methodnamei{} contribute to a higher time cost for solving individual instances. However, this property leads to better \methodnamei{}'s capacity to generate appropriate subgoals based on the current external environment compared to \methodnameo{}.

For example, when tackling a complex question ``\texttt{What government position was held by the woman who portrayed Corliss Archer in the film Kiss and Tell?}'', \methodnamei{} can first identify the woman who portrayed Corliss Archer in Kiss and Tell, Shirley Temple, then ask the government position she held. However, though \methodnameo{} can first ask who the woman is as well, without the interaction with external knowledge in Wikipedia, it will then generate a subgoal which inquires the government position held by Madonna, a random entity irrelevant with the original question. Hence, \methodnameo{} is a more efficient solution, but not as effective as \methodnamei{} due to the lack of the adaptability to external environments.

We also notice that in Tab.~\ref{table:exp-maths}, \methodnameo{} achieves superior performance on math tasks. Consider a mathematical problem such as, ``\texttt{James decides to run 3 sprints 3 times a week. He runs 60 meters each sprint. How many total meters does he run a week?}'' Even if the agent calculated how many times James sprints a week, which is $3\times3=9$, the mere number 9 does not affect the next subgoal generation. This is because no matter the result is 9 or 10000, the next high-level subgoal to calculate the total meters remains the same for solving the question. Therefore, the high environment adaptability of \methodnamei{} may not be a good supplement on math tasks.

\section{More Unified Training Results}
\label{app:unified}

\begin{table}[h]
\centering
\scalebox{0.68}{
\begin{tabular}{ccccc}
\toprule
\textbf{Agents} & \textbf{Web}                  & \textbf{QA}   & \textbf{Multimodal} & \textbf{Math} \\ \midrule
\methodnamei{}$_{\mathrm{X}}$-13B &     31.3       & 65.3/50.2/31.4   & 72.4/58.2 & \textbf{51.9}/\textbf{66.3} \\
\methodnamei{}$_{\mathrm{All}}$-13B &    \textbf{31.9}    &  \textbf{66.7}/\textbf{51.0}/\textbf{31.6}  & \textbf{72.8}/58.2 & 50.5/65.8\\
\bottomrule
\end{tabular}
}
\caption{Unified training performance on trained task types. We aggregate the performance on held-in/held-out datasets for each trained task type with the symbol `/'. As discussed in the footnote 2 of \S\ref{sec:exp-generalizability}, \methodnameo{} is not applicable to web tasks. Thus, we only conduct unified training for \methodnamei{} as it can be universally applied to any task types.}
\label{table:appendix-unified_training}
\end{table}

We evaluate the performance of \methodnamei{}-13B after the unified training that leverages combined annotations from four distinct training task types. We observe that the unified training enhances performance across a majority of tasks, including web, QA, and multimodal tasks. The decline in performance for mathematical tasks is marginal, which is only 0.7\% and 1.4\%. The unified task representation may enable \methodname{} agents to uncover intrinsic links among these complex interactive tasks, thereby yielding additional advantages for certain training tasks.

\section{Execution Tools Associated with Action Interfaces}
\label{app:tools}

\begin{table*}[t]
\centering
\vspace{-0em}
\begin{subtable}[t]{\textwidth}
\centering
\scalebox{0.7}{
\begin{tabular}{cccc}
\toprule
\textbf{Task Type}                  & \textbf{Action Types}   & \textbf{Function Descriptions} & \textbf{Tools} \\ \midrule
\multirow{7}{*}{QA} & \texttt{KnowledgeQuery(Entity) -> Knowledge}    & Query the entity knowledge   & Wikipedia, Google Search              \\ \cmidrule{2-4}
& \makecell{\texttt{ParagraphRetrieval(Knowledge, Query)} \\ \texttt{-> Paragraphs}}    & \makecell{Retrieve relevant paragraphs \\ according to the query}   & \texttt{dpr-reader-multiset-base}              \\ \cmidrule{2-4}
& \texttt{QA(Context, Query) -> Answer}    & \makecell{Answer the query based on \\ the given context}   & GPT-series/open LLMs              \\ \cmidrule{2-4}
& \texttt{Calculator(Expression) -> Value}
    & Calculate given math expressions   & WolframAlpha \\ \bottomrule
\end{tabular}
}
\caption{Actions used in complex QA tasks.}
\label{table:appendix-tools-qa}
\end{subtable}
\vspace{9pt}

\begin{subtable}[t]{\textwidth}
\centering
\scalebox{0.69}{
\begin{tabular}{cccc}
\toprule
\textbf{Task Type}                  & \textbf{Action Types}   & \textbf{Function Descriptions} & \textbf{Implementation} \\ \midrule
\multirow{6}{*}{Web} & \texttt{Click(Env, Query) -> Tag}    & \makecell{Locate the tag to be clicked according to the query}   & \multirow{6}{*}{\makecell{HTML Simulator}}              \\ \cmidrule{2-3}
& \texttt{Type(Env, Query, Text) -> Tag, Text}    & \makecell{Locate the relevant tag according to the query \\ and output the typed text}    &               \\ \cmidrule{2-3}
& \texttt{Select(Env, Query, Text) -> Tag, Text}    & \makecell{Locate the relevant tag according to the query \\ and output the selected option}   &               \\
\bottomrule
\end{tabular}
}
\caption{Actions used in web tasks.}
\label{table:appendix-tools-web}
\end{subtable}
\vspace{9pt}

\begin{subtable}[t]{\textwidth}
\centering
\scalebox{0.7}{
\begin{tabular}{cccc}
\toprule
\textbf{Task Type}                  & \textbf{Action Types}   & \textbf{Function Descriptions} & \textbf{Implementation} \\ \midrule
\multirow{8}{*}{Math}      & \texttt{Calculator(Expression) -> Value}
& Calculate given math expressions   & \multirow{6}{*}{WolframAlpha}     \\ \cmidrule{2-3}
& \texttt{SetEquation(Expression) -> Equation} & Set equations based on given expressions &                    \\ \cmidrule{2-3}
& \texttt{SolveEquation(Equation) -> Solutions}      & Solve the set equations  &                  \\ \cmidrule{2-3}
& \texttt{Define(Variable) -> Variable} & Define a variable &                    \\ \cmidrule{2-3}
& \texttt{SolveInequality(Inequality) -> Solutions}      & Solve the given inequality  &                  \\ \cmidrule{2-4}
& \texttt{Code(Function\_Description) -> Code} & Generate codes for math functions & \texttt{gpt-3.5-turbo}        \\ \cmidrule{2-4}
& \texttt{Count(List) -> Number}      & Count the element number in a list  & Python                 \\ \bottomrule
\end{tabular}
}
\caption{Actions used in math tasks.}
\label{table:appendix-tools-math}
\end{subtable}
\vspace{9pt}

\begin{subtable}[t]{\textwidth}
\centering
\scalebox{0.7}{
\begin{tabular}{cccc}
\toprule
\textbf{Task Type}                  & \textbf{Action Types}   & \textbf{Function Descriptions} & \textbf{Implementation} \\ \midrule
\multirow{3}{*}{Multimodal} & \texttt{QA(Context, Query) -> Answer}    & \makecell{Answer the query based on \\ the given context}   & \texttt{LLAMA-2-13B-chat}              \\ \cmidrule{2-4}
& \texttt{VQA(Image\_Context, Query) -> Answer}    & \makecell{Answer the query based on \\ the given image context}   & \texttt{LLAVA-1.5-7B}              \\ \bottomrule
\end{tabular}
}
\caption{Actions used in multimodal tasks.}
\label{table:appendix-tools-multimodal}
\end{subtable}

\caption{Action interfaces and execution module implementations for complex interactive tasks.}
\label{table:appendix-tools}
\end{table*}

For each available action defined in the action interfaces, there is at least one corresponding backend execution tool to facilitate the implementation of the concrete actions.

Displayed in Tab.~\ref{table:appendix-tools-qa}, for QA tasks, we rely on Wikipedia and Google search APIs to locate relevant entity knowledge. Besides, we leverage a semantic matching model \texttt{dpr-reader-multiset-base}\footnote{\url{https://huggingface.co/facebook/dpr-reader-multiset-base}.} used in Dense Passage Retrieval (DPR)~\citep{karpukhin-etal-2020-dense} to capture paragraphs according to the query. Following ReWOO~\citep{xu2023rewoo}, we also include GPT-series model as a simple QA tool to answer the query based on our retrieved knowledge or previously interaction history.

Demonstrated in Tab.~\ref{table:appendix-tools-web}, for web tasks, the actions are real mouse and keyboard operations including typing, clicking and selecting HTML tags. To locate the relevant HTML tags to be operated, following AgentBench evaluation, we use a DeBERTa model\footnote{\url{https://huggingface.co/osunlp/MindAct_CandidateGeneration_deberta-v3-base}.} that ranks and retrieves the tags according to the current action.

As shown in Tab.~\ref{table:appendix-tools-math}, for math tasks, the main execution tool is WolframAlpha API~\footnote{\url{https://www.wolframalpha.com/}.} as it can perform a large collection of mathematical functions such as calculating formulas and solving equations. For complex math operations such as sorting, we leverage OpenAI Codex~\citep{chen2021evaluating} to generate a short code snippet for the execution.

For multimodal tasks, as illustrated in Tab. \ref{table:appendix-tools-multimodal}, both Visual Question Answering (VQA) and Question Answering (QA) tools are considered. The employed VQA model  is LLAVA-1.5-7B~\citep{liu2023improved}, while the utilized QA model is LLAMA-2-13B-chat~\citep{touvron2023llama2}.

For the unseen task WebShop, the actions include \texttt{Search}, \texttt{FeatureRetrieve}, \texttt{Pick}, and \texttt{Click}. The implementation of \texttt{Search} and \texttt{Click} relies on the embedded implementation already provided in official WebShop virtual environment\footnote{\url{https://github.com/princeton-nlp/WebShop}.}. \texttt{FeatureRetrieve} and \texttt{Pick} are based on \texttt{dpr-reader-multiset-base}, which helps to select the most relevant items and their item features according to the query.

For the unseen task InterCode$_{\mathrm{SQL}}$, the action interfaces include all the possible commands and functions provided in SQL programming language.

\section{Details of Performance Evaluation}
\label{app:eval}
\paragraph{Metrics.} 
Here we mainly discuss the special metrics adopted to evaluate the agent performance. For HotpotQA, instead of merely using strict exact matching, we follow \citet{xu2023rewoo} to also use GPT-4 as an evaluator to judge whether the predicted answer shares the same semantics with the gold answer. We call this metric as LLM accuracy, frequently mentioned in \S\ref{sec:exp}. For Mind2Web, we adopt the same metric step success rate used for AgentBench evaluation. A step is deemed successful solely when both the chosen HTML tag and predicted action type exactly match the gold action. For WebShop, we leverage the reward utilized in both AgentBench and original WebShop paper, which quantify the similarity between gold and predicted products with regard to the product titles and selected attributes. 

\paragraph{Evaluation Data.}
Following \citet{xu2023rewoo}, we only evaluate 300 and 1000 randomly selected examples from StrategyQA and HotpotQA evaluation set, respectively. The results reported in Tab.~\ref{table:exp-complex-qa} are the average performance on three different sets of sampled data. Regarding Mind2Web, we only evaluate on the ``cross-domain'' test set that AgentBench utilizes for evaluation. For WebShop, we evaluate the first 500 instances from the entire test set as AgentBench used to do. We leverage the entire evaluation set of the other testing benchmarks for assessment.

\onecolumn
\section{In-Context Examples in Conversion Prompts}
\label{app:conversion-prompts}
As discussed in \S\ref{sec:annotations_conversion_prompts}, in-context examples are helpful to instruct LLMs to generate annotations in our expected format. For each training task types, we showcase one in-context example to help readers better understand how the prompting conversion method works and the format of our expected annotations. We highlight subgoals, their actions and execution results with \colorbox[rgb]{1, 1, 0.8}{yellow}, \colorbox[rgb]{1, 0.93, 0.93}{red} and \hlcyan{blue}, respectively.

\subsection{In-Context Example For Obtaining Math Task Annotations}
\noindent
\colorbox[rgb]{ .95,  .95,  .95}{
\begin{minipage}{\textwidth}
\small
\texttt{Please convert natural language plans into a series of subgoals and their corresponding actions that lead to the successful implementation with respect to the given instructions. Please use `R[number]' to represent the intermediate results for each subgoal, without generating any exact values. Please also use functions to represent the corresponding actions. For the actions, they must be one of `Calculator', `SetEquation', `SolveEquation', `SolveInequality', `Count', `Code', and `Define'.}

\hspace*{\fill}

\texttt{Example 1:}

\hspace*{\fill}

\texttt{Task: Peter goes to the store to buy a soda. The soda costs \$.25 an ounch. He brought \$2 with him and leaves with \$.50. How many ounces of soda did he buy?}

\hspace*{\fill}

\texttt{Natural language plan:}

\texttt{He spend \$1.5 on soda because 2 - .5 = 1.5 He bought 6 ounces of soda because 1.5 / .25 = 6}

\hspace*{\fill}

\texttt{Subgoal-based plan:}

\texttt{\colorbox[rgb]{1, 1, 0.8}{Subgoal 1: Calculate how much the soda costs in total.}}

\texttt{Action 1-1: \colorbox[rgb]{1, 0.93, 0.93}{R1 = Calculator(2 - 0.5)} = \hlcyan{1.5}}

\hspace*{\fill}

\texttt{\colorbox[rgb]{1, 1, 0.8}{Subgoal 2: Calculate the ounces of soda the price per ounch.}}

\texttt{Action 2-1: \colorbox[rgb]{1, 0.93, 0.93}{R2 = Calculator(R1 / 0.25)} = \hlcyan{6}}
\end{minipage}
}

\subsection{In-Context Example For Obtaining Complex QA Task Annotations}
\noindent
\colorbox[rgb]{ .95,  .95,  .95}{\begin{minipage}{\textwidth}
\small
\texttt{Please convert natural language plans into a series of subgoals and their corresponding actions that lead to the successful implementation with respect to the given instructions. Please use `R[number]' to represent the intermediate results for each subgoal, without generating any exact values. Please also use functions to represent the corresponding actions. For the actions, they must be one of one of `KnowledgeQuery', `ParagraphRetrieve', `QA', `Calculator' and `Code'.}

\hspace*{\fill}

\texttt{Example 1:}

\hspace*{\fill}

\texttt{Task: Are more people today related to Genghis Khan than Julius Caesar?}

\hspace*{\fill}

\texttt{Natural language plan:}

\texttt{We find relevant facts: Julius Caesar had three children. Genghis Khan had sixteen children. Modern geneticists have determined that out of every 200 men today has DNA that can be traced to Genghis Khan. We need to answer these questions: 1. How many kids did Julius Caesar have? (Can be answered based on paragraph `Julius Caesar-75') 2. How many kids did Genghis Khan have? (Can be answered based on paragraph `Genghis Khan-17') 3. Is \#2 greater than \#1? Based on these evidences and decomposed questions, the answer is True.}

\hspace*{\fill}

\texttt{Subgoal-based plan:}

\texttt{\colorbox[rgb]{1, 1, 0.8}{Subgoal 1: Obtain the number of the kids that Julius Caesar had.}}

\texttt{Action 1-1: \colorbox[rgb]{1, 0.93, 0.93}{R1 = KnowledgeQuery(Julius Caesar)} = WikipediaPage(Julius Caesar)}

\texttt{Action 1-2: \colorbox[rgb]{1, 0.93, 0.93}{R2 = ParagraphRetrieve(R1, Query: How many kids did Julius Caesar have?)} = Paragraph(Julius Caesar-75).}

\texttt{Action 1-3: \colorbox[rgb]{1, 0.93, 0.93}{R3 = QA([R2], Question: How many kids did Julius Caesar have?)} = \hlcyan{3}.}

\hspace*{\fill}

\texttt{\colorbox[rgb]{1, 1, 0.8}{Subgoal 2: Obtain the number of the kids that Genghis Khan had.}}

\texttt{Action 2-1: \colorbox[rgb]{1, 0.93, 0.93}{R4 = KnowledgeQuery(Genghis Khan)} = WikipediaPage(Genghis Khan).}

\texttt{Action 2-2: \colorbox[rgb]{1, 0.93, 0.93}{R5 = ParagraphRetrieve(R4, Query: How many kids did Genghis Khan have?)} = Paragraph(Genghis Khan-17).}

\texttt{Action 2-3: \colorbox[rgb]{1, 0.93, 0.93}{R6 = QA([R5], Question: How many kids did Genghis Khan have?)} = \hlcyan{16}.}

\hspace*{\fill}

\texttt{\colorbox[rgb]{1, 1, 0.8}{Subgoal 3: Determine if Genghis Khan had more kids.}}

\texttt{Action 3-1: \colorbox[rgb]{1, 0.93, 0.93}{R7 = Calculator(R6 > R3)} = \hlcyan{True}}
\end{minipage}}

\clearpage

\subsection{In-Context Example For Obtaining Web Task Annotations}
Since the data source for converting annotations, Mind2Web, already provides the ground-truth execution results after each action, as discussed in \S\ref{sec:annotations_conversion_prompts}, we do not ask LLMs to capture each action's execution results. Therefore, there are no parts highlighted with blue in the in-context example.
\vspace{1em}

\noindent \colorbox[rgb]{ .95,  .95,  .95}{\begin{minipage}{\textwidth}
\small
\texttt{Please convert natural language plans into a series of subgoals and their corresponding actions that lead to the successful implementation with respect to the given instructions. Please use `R[number]' to represent the intermediate results for each subgoal, without generating any exact values. Please also use functions to represent the corresponding actions. For the actions, they must be one of they must be one of `TYPE', `CLICK', and `SELECT'.}

\hspace*{\fill}

\texttt{Example 1:}

\hspace*{\fill}

\texttt{Task: Find a Ricky Kej track to listen and share which has been added in the last year and is between 2 to 10 minutes.}

\hspace*{\fill}

\texttt{Natural language plan:}

\texttt{[searchbox]  Search $\longrightarrow$ TYPE: Ricky Kej; [link]  Search for ``Ricky Kej'' $\longrightarrow$ CLICK; [link]  Tracks $\longrightarrow$ CLICK; [link]  Added any time $\longrightarrow$ CLICK; [link]  Past year $\longrightarrow$ SELECT; [link]  Any length $\longrightarrow$ CLICK; [link]  2-10 min $\longrightarrow$ CLICK; [link]  To listen to $\longrightarrow$ CLICK; [link]  To share $\longrightarrow$ CLICK}

\hspace*{\fill}

\texttt{Subgoal-based plan:}

\texttt{\colorbox[rgb]{1, 1, 0.8}{Subgoal 1: Type Ricky Kej to search his songs.}}

\texttt{Action 1-1: \colorbox[rgb]{1, 0.93, 0.93}{R1 = TYPE(Env, QUERY: Type Ricky Kej to search his songs, TEXT: Ricky Kej)}}

\hspace*{\fill}

\texttt{\colorbox[rgb]{1, 1, 0.8}{Subgoal 2: Click on the option to search for Ricky Rej.}}

\texttt{Action 2-1: \colorbox[rgb]{1, 0.93, 0.93}{R2 = CLICK(R1, QUERY: Click on the option to search for Ricky Rej)}}

\hspace*{\fill}

\texttt{\colorbox[rgb]{1, 1, 0.8}{Subgoal 3: Choose tracks as the search category.}}

\texttt{Action 3-1: \colorbox[rgb]{1, 0.93, 0.93}{R3 = CLICK(R2, QUERY: Choose tracks as the search category)}}

\hspace*{\fill}

\texttt{\colorbox[rgb]{1, 1, 0.8}{Subgoal 4: Find the region to adjust the added time of our interested track.}}

\texttt{Action 4-1: \colorbox[rgb]{1, 0.93, 0.93}{R4 = CLICK(R3, QUERY: Find the region to adjust the added time of our interested track)}}

\hspace*{\fill}

\texttt{\colorbox[rgb]{1, 1, 0.8}{Subgoal 5: Choose the last year as the added date.}}

\texttt{Action 5-1: \colorbox[rgb]{1, 0.93, 0.93}{R5 = SELECT(R4, QUERY: Choose the last year as the added date, TEXT: Past year)}}

\hspace*{\fill}

\texttt{\colorbox[rgb]{1, 1, 0.8}{Subgoal 6: Find the region to adjust the track length of our interested track.}}

\texttt{Action 6-1: \colorbox[rgb]{1, 0.93, 0.93}{R6 = CLICK(R5, QUERY: Find the region to adjust the track length of our interested track)}}

\hspace*{\fill}

\texttt{\colorbox[rgb]{1, 1, 0.8}{Subgoal 7: Choose 2 to 10 minutes as the track length.}}

\texttt{Action 7-1: \colorbox[rgb]{1, 0.93, 0.93}{R7 = CLICK(R6, QUERY: Choose 2 to 10 minutes as the track length)}}

\hspace*{\fill}

\texttt{\colorbox[rgb]{1, 1, 0.8}{Subgoal 8: Listen to our searched track.}}

\texttt{Action 8-1: \colorbox[rgb]{1, 0.93, 0.93}{R8 = CLICK(R7, QUERY: Listen to our searched track)}}

\hspace*{\fill}

\texttt{\colorbox[rgb]{1, 1, 0.8}{Subgoal 9: Share our searched track.}}

\texttt{Action 9-1: \colorbox[rgb]{1, 0.93, 0.93}{R9 = CLICK(R8, QUERY: Share our searched track)}}
\end{minipage}}

\clearpage

\subsection{In-Context Example For Obtaining Multimodal Task Annotations}
\noindent
\colorbox[rgb]{ .95,  .95,  .95}{
\begin{minipage}{\textwidth}
\small
\texttt{Please convert natural language plans into a series of subgoals, their corresponding actions that lead to the successful implementation with respect to the given instructions. When generating the actions, please also attach the action's results contained in the natural language plans. Please use 'R[number]' to represent the execution results for each action. Please also use functions to represent the corresponding actions. For the actions, they must be one of the available actions, 'QA', 'VQA'.}

\hspace*{\fill}

\texttt{Example 1:}

\hspace*{\fill}

\texttt{Task: If the cameraman were driving what do they have to do from this position? There're some choices: A. turn left, B. drive straight, C. reverse course, D. turn right.}

\hspace*{\fill}

\texttt{Natural language plan:}

\texttt{The would have to turn right because the lane has right turn arrows painted on it. The arrow on the street indicates that this lane can only go in one direction at the intersection. The sign on the road says to turn right. Overall, the final answer is 'turn right'.}

\hspace*{\fill}

\texttt{Subgoal-based plan:}

\texttt{\colorbox[rgb]{1, 1, 0.8}{Subgoal 1: Describe the shape of the sign on the road lane the cameraman is in from the image.}}

\texttt{Action 1-1: \colorbox[rgb]{1, 0.93, 0.93}{R1 = VQA([IMG], Question: What's the sign on the road lane?)} = \hlcyan{There's a right turn arrow on the road.}}

\hspace*{\fill}

\texttt{\colorbox[rgb]{1, 1, 0.8}{Subgoal 2: Answer which lane the cameraman is in.}}

\texttt{Action 2-1: \colorbox[rgb]{1, 0.93, 0.93}{R2 = QA([R1], Question: Which lane is the cameraman in?)} = \hlcyan{The cameraman is in right turn lane.}}

\hspace*{\fill}

\texttt{\colorbox[rgb]{1, 1, 0.8}{Subgoal 3: Answer what the cameraman has to do from his position.}}

\texttt{Action 3-1: \colorbox[rgb]{1, 0.93, 0.93}{R3 = QA([R1, R2], Question: What does the cameraman have to do from his position?)} = \hlcyan{turn right.}}

\hspace*{\fill}

\texttt{\colorbox[rgb]{1, 1, 0.8}{Subgoal 4: Select the choice from the answer choices.}}

\texttt{Action 4-1: \colorbox[rgb]{1, 0.93, 0.93}{R4 = VQA([IMG], Question: If the cameraman were driving what do they have to do from this} \\ \colorbox[rgb]{1, 0.93, 0.93}{position? We can also see that There's a right turn arrow on the road, The cameraman is in right turn} \\ \colorbox[rgb]{1, 0.93, 0.93}{lane, turn right. There're some choices: A. turn left, B. drive straight, C. reverse course, D. turn} \\ \colorbox[rgb]{1, 0.93, 0.93}{right. You must choose one from 'A', 'B', 'C' or 'D'. Output only one letter!)} = \hlcyan{D}}
\end{minipage}
}

\clearpage

\section{In-Context Example for Adapting Modules to Unseen Task --- WebShop}
\label{app:webshop}
As discussed in \S\ref{sec:exp-generalizability}, the in-context examples are used for helping modules understand how to plan and ground to novel available actions. Mentioned in App.~\ref{app:tools}, WebShop involves four unseen actions in training annotations: \texttt{Search}, \texttt{FeatureRetrieve}, \texttt{Pick}, and \texttt{Click}. The definition of \texttt{Click} is different from the one defined for web task. In WebShop task, the argument of \texttt{Click} is the item index. However, \texttt{Click}'s input arguments in Mind2Web include an HTML environment description and a specific query regarding the goal of click action. We showcase one of the two in-context WebShop-related examples used for the inference of planning and grounding modules, respectively. 

\subsection{In-Context WebShop-Related Example for Planning Module}
\vspace{6pt}
\noindent
\colorbox[rgb]{ .95,  .95,  .95}{\begin{minipage}{\textwidth}
\small
\texttt{<|user|>}

\texttt{Please provide a reasonable subgoal-based plan to solve the given task.}

\texttt{Task: Find me slim fit, loose fit women's tops, tees \& blouses with long sleeve, short sleeve with color: a01\#black, and size: large, and price lower than 50.00 dollars}

\hspace*{\fill}

\texttt{<|assistant|>}

\texttt{Subgoal 1: Search two relevant items and the product names according to the instruction.}

\hspace*{\fill}

\texttt{<|user|>}

\texttt{The execution result for Subgoal 1 is b09s3bn15c - Mens Linen Shirt,Men's Striped Shirts Casual Short Sleeve Button Down Shirts Regular Fit Hawaiian Shirts Beach Tees Tops
** b094q7b3ss - Women Cold Shoulder Tops, Summer Butterfly Print Shirts Fashion Casual Short Sleeve Plus-Size Tunic Top Tee and Blouse.}

\hspace*{\fill}

\texttt{<|assistant|>}

\texttt{Subgoal 2: Select the most relevant features of item b09s3bn15c.}

\hspace*{\fill}

\texttt{<|user|>}

\texttt{The execution result for Subgoal 2 is short, casual, shoulder.}

\hspace*{\fill}

\texttt{<|assistant|>}

\texttt{Subgoal 3: Select the most relevant features of item b094q7b3ss.}

\hspace*{\fill}

\texttt{<|user|>}

\texttt{The execution result for Subgoal 3 is black, large, x-large.}

\hspace*{\fill}

\texttt{<|assistant|>}

\texttt{Subgoal 4: Pick up the most related one from the two relevant items according to the product names and their features.}

\hspace*{\fill}

\texttt{<|user|>}

\texttt{The execution result for Subgoal 4 is b094q7b3ss.}

\hspace*{\fill}

\texttt{<|assistant|>}

\texttt{Subgoal 5: Buy this item by selecting its relevant features.}

\hspace*{\fill}

\texttt{<|user|>}

\texttt{The execution result for Subgoal 5 is Done.}
\end{minipage}}

\clearpage

\subsection{In-Context WebShop-Related Example for Grounding Module}
\vspace{6pt}
\colorbox[rgb]{ .95,  .95,  .95}{\begin{minipage}{\textwidth}
\small
\texttt{<|user|>}

\texttt{Please ground the given subgoal to corresponding executable actions for solving the given task. The grounded actions must be the one in available action list.}

\hspace*{\fill}

\texttt{The available action list is `Search', `FeatureRetrieve', `Pick' and `Click'. Search(Query): Search related items according to the Query; FeatureRetrieve(Feature\_list, Query): Select the most relevant features from Feature\_list according to Query; Pick(Item\_names, Item\_features, Query): Pick the most relevant item from Item\_names according to Query, Item\_names and Item\_features; Click(Item): Click the Item to check more information.}

\hspace*{\fill}

\texttt{Task: Find me slim fit, loose fit women's tops, tees \& blouses with long sleeve, short sleeve with color: a01\#black, and size: large, and price lower than 50.00 dollars}

\texttt{Subgoal to be grounded: Subgoal 1: Search two relevant items and the product names according to the instruction.}

\hspace*{\fill}

\texttt{<|assistant|>}

\texttt{R1 = Search(Find me slim fit, loose fit women's tops, tees \& blouses with long sleeve, short sleeve with color: a01\#black, and size: large, and price lower than 50.00
dollars)}

\hspace*{\fill}

\texttt{<|user|>}

\texttt{Subgoal to be grounded: Subgoal 2: Select the most relevant features of item b09s3bn15c.}

\hspace*{\fill}

\texttt{<|assistant|>}

\texttt{R2 = Click(b09s3bn15c); R3 = FeatureRetrieve(R2, Query: Find me slim fit, loose fit women's tops, tees \& blouses with long sleeve, short sleeve with color: a01\#black, and size: large, and price lower than 50.00 dollars)}

\hspace*{\fill}

\texttt{<|user|>}

\texttt{Subgoal to be grounded: Subgoal 3: Select the most relevant features of item b094q7b3ss.}

\hspace*{\fill}

\texttt{<|assistant|>}

\texttt{R4 = Click(< prev); R5 = Click(b094q7b3ss); R6 = FeatureRetrieve(R5, Query: Find me slim fit, loose fit women's tops, tees \& blouses with long sleeve, short sleeve with color: a01\#black, and size: large, and price lower than 50.00 dollars)}

\hspace*{\fill}

\texttt{<|user|>}

\texttt{Subgoal to be grounded: Subgoal 4: Pick up the most related one from the two relevant items according to the product names and their features.}

\hspace*{\fill}

\texttt{<|assistant|>}

\texttt{R7 = Pick(Item\_names: R1, Item\_features: [R3, R6], Query: Find me slim fit, loose fit women's tops, tees \& blouses with long sleeve, short sleeve with color: a01\#black,
and size: large, and price lower than 50.00 dollars)}

\hspace*{\fill}

\texttt{<|user|>}

\texttt{Subgoal to be grounded: Subgoal 5: Buy this item by selecting its relevant features.}

\hspace*{\fill}

\texttt{<|assistant|>}

\texttt{R8 = Click(< prev); R9 = Click(R7); R10 = Click(black); R11 = Click(large); R12 = Click(x-large); R13 = Click(buy now)}
\end{minipage}}

\clearpage

\section{In-Context Example for Adapting Modules to Unseen Task --- InterCode$_\mathrm{SQL}$}
\label{app:intercode}
Similar to \S \ref{app:webshop}, as InterCode$_\mathrm{SQL}$ is another unseen task, we showcase one of the three in-context InterCode$_\mathrm{SQL}$-related examples used for the inference of planning and grounding modules, respectively.

\subsection{In-Context InterCode$_\mathrm{SQL}$-Related Example for Planning Module}
\vspace{6pt}
\noindent
\colorbox[rgb]{ .95,  .95,  .95}{\begin{minipage}{\textwidth}
\small
\texttt{<|user|>}

\texttt{Please provide a reasonable subgoal-based plan to solve the given task.}

\texttt{Task: What are the names of poker players, ordered ascending by the number of final tables they have made?}

\hspace*{\fill}

\texttt{<|assistant|>}

\texttt{Subgoal 1: Find all the related tables.}

\hspace*{\fill}

\texttt{<|user|>}

\texttt{The execution result for Subgoal 1 is Results: Table players has Name (type: TEXT), People\_ID (type: INT); Table poker\_player has Final\_Table\_Made (type: INT), People\_ID (type: INT).; Subgoal 2: Write the selection objects: the names of the people who are also poker players.}

\hspace*{\fill}

\texttt{<|assistant|>}

\texttt{Subgoal 2: Write the selection objects: the names of the people who are also poker players.}

\hspace*{\fill}

\texttt{<|user|>}

\texttt{The execution result for Subgoal 2 is T1.Name FROM people AS T1 JOIN poker\_player AS T2 ON T1.People\_ID = T2.People\_ID.}

\hspace*{\fill}

\texttt{<|assistant|>}

\texttt{Subgoal 3: Write the selection condition: sort all the poker players by the number of final tables they have made in ascending order.}

\hspace*{\fill}

\texttt{<|user|>}

\texttt{The execution result for Subgoal 3 is ORDER BY T2.Final\_Table\_Made.}

\hspace*{\fill}

\texttt{<|assistant|>}

\texttt{Subgoal 4: Finish the SQL query to the names of poker players, ordered ascending by the number of final tables they have made.}

\hspace*{\fill}

\texttt{<|user|>}

\texttt{The execution result for Subgoal 4 is SELECT T1.Name FROM people AS T1 JOIN poker\_player AS T2 ON T1.People\_ID = T2.People\_ID ORDER BY T2.Final\_Table\_Made.}
\end{minipage}}

\clearpage

\subsection{In-Context InterCode$_\mathrm{SQL}$-Related Example for Grounding Module}
\vspace{6pt}
\colorbox[rgb]{ .95,  .95,  .95}{\begin{minipage}{\textwidth}
\small
\texttt{<|user|>}

\texttt{Please ground the given subgoal to corresponding executable actions for solving the given task. The grounded actions must be the one in available action list.}

\hspace*{\fill}

\texttt{Task: What are the names of poker players, ordered ascending by the number of final tables they have made?}

\texttt{Subgoal to be grounded: Subgoal 1: Find all the related tables.}

\hspace*{\fill}

\texttt{<|assistant|>}

\texttt{R1 = SHOW TABLES}

\hspace*{\fill}

\texttt{<|user|>}

\texttt{Subgoal to be grounded: Subgoal 2: Write the selection objects: the names of the people who are also poker players.}

\hspace*{\fill}

\texttt{<|assistant|>}

\texttt{R2 = T1.Name FROM people AS T1 JOIN poker\_player AS T2 ON T1.People\_ID = T2.People\_ID}

\hspace*{\fill}

\texttt{<|user|>}

\texttt{Subgoal to be grounded: Subgoal 3: Write the selection condition: sort all the poker players by the number of final tables they have made in ascending order.}

\hspace*{\fill}

\texttt{<|assistant|>}

\texttt{R3 = ORDER BY T2.Final\_Table\_Made}

\hspace*{\fill}

\texttt{<|user|>}

\texttt{Subgoal to be grounded: Subgoal 4: Finish the SQL query to the names of poker players, ordered ascending by the number of final tables they have made.}

\hspace*{\fill}

\texttt{<|assistant|>}

\texttt{R4 = SELECT T1.Name FROM people AS T1 JOIN poker\_player AS T2 ON T1.People\_ID = T2.People\_ID ORDER BY T2.Final\_Table\_Made}
\end{minipage}}

\clearpage

\section{Multimodal Task Case Study}
We provide two cases in the multimodal datasets, A-OKVQA and ScienceQA, to show the effectiveness of \methodname{}. For the A-OKVQA case, \methodname{} first identifies the device brand with the aid of VQA tool. After \methodname{} planning module knows that the device is a Nintendo Wii controller, the module would generate the next subgoal ``\texttt{Answer the country of origin for the Nintendo Wii controller}''. Finally, \methodname{} selects the closest option from the given answer choices. For the ScienceQA case, \methodname{} identifies the highlighted area on the map, which includes Middle East area. Then the planning and grounding modules would leverage QA tool to answer which continent Middle East is located at, and then pick up the correct answer ``\texttt{Asia}''.

\begin{figure*}[htbp]
\centering
\includegraphics[width=\textwidth, trim=0 70 0 0, clip]{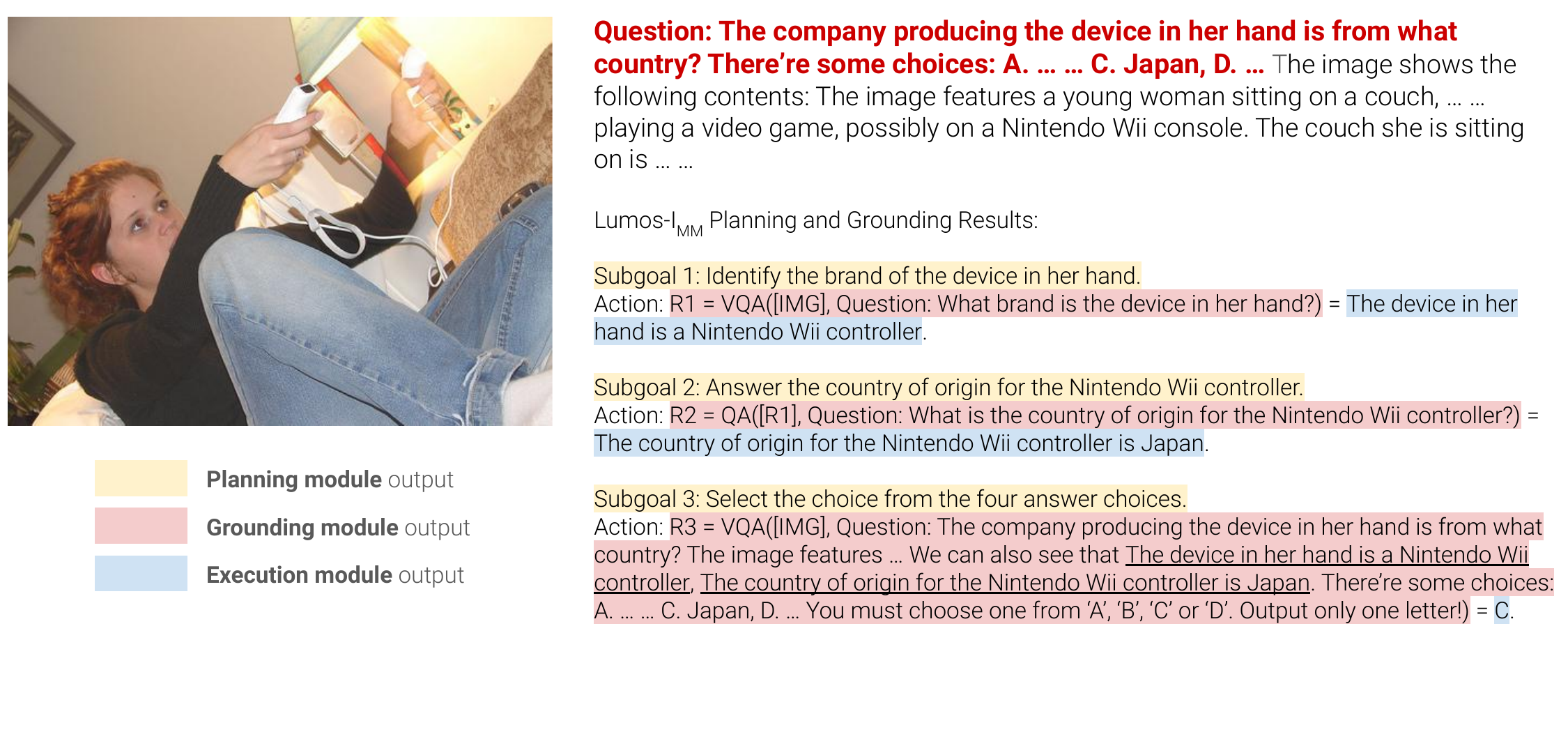}
\caption{\methodnamei{}$_\mathrm{MM}$ case study on A-OKVQA.}
\label{fig:multimodal-case-1}
\end{figure*}

\begin{figure*}[htbp]
\centering
\includegraphics[width=\textwidth, trim=0 90 30 0, clip]{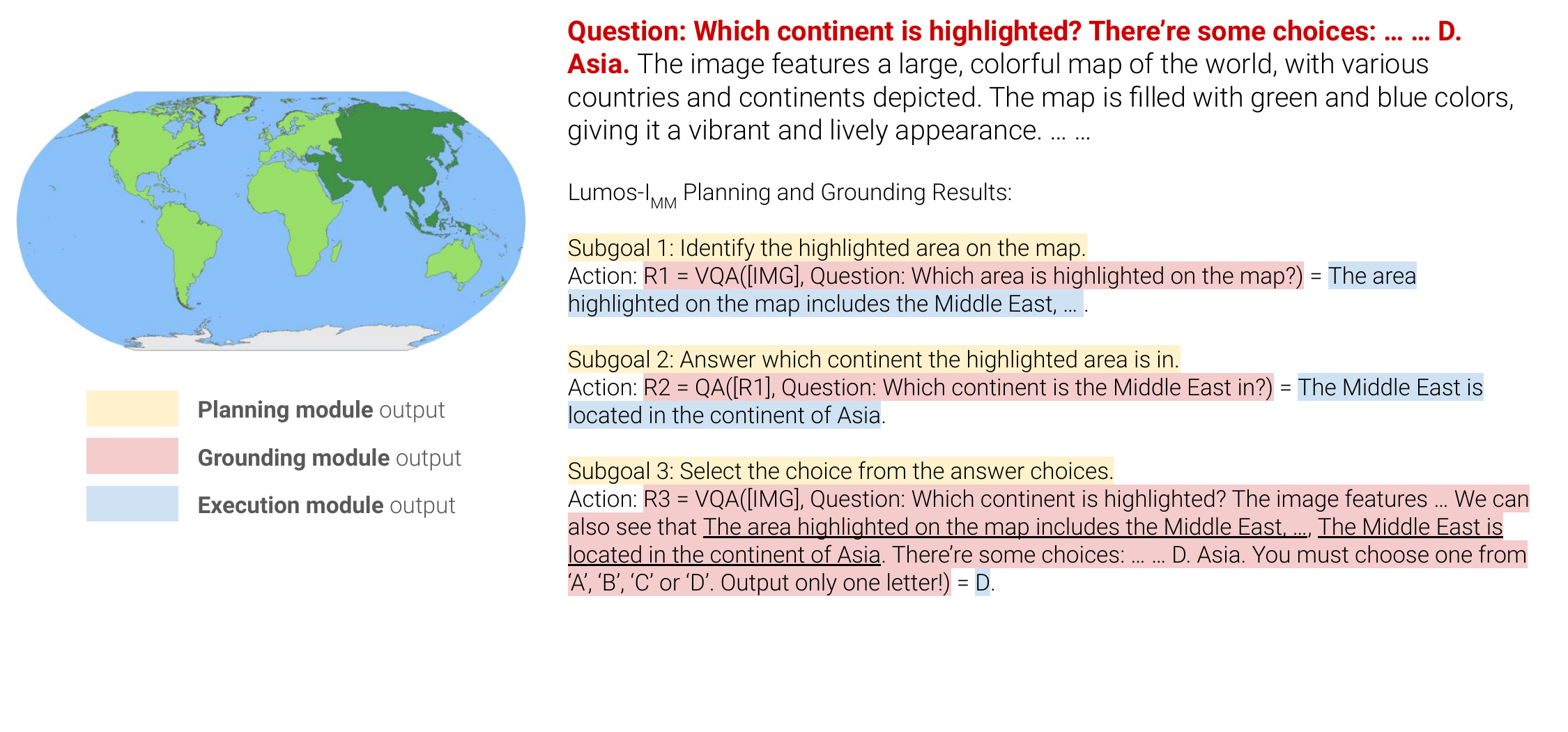}
\caption{\methodnamei{}$_\mathrm{MM}$ case study on ScienceQA.}
\label{fig:multimodal-case-2}
\end{figure*}

\end{document}